\documentclass[lettersize,journal]{IEEEtran}
\usepackage{amsmath,amsfonts}
\usepackage{algorithmic}
\usepackage{algorithm}
\usepackage{array}
\usepackage[caption=false,font=normalsize,labelfont=sf,textfont=sf]{subfig}
\usepackage{textcomp}
\usepackage{stfloats}
\usepackage{url}
\usepackage{verbatim}
\usepackage{graphicx}
\usepackage{cite}
\usepackage{color}
\usepackage{times}
\usepackage{soul}
\usepackage{url}
\usepackage[hidelinks]{hyperref}
\usepackage[utf8]{inputenc}
\usepackage[small]{caption}
\usepackage{graphicx}
\usepackage{amsmath}
\usepackage{amsthm}
\usepackage{booktabs}
\usepackage{algorithm}
\usepackage{algorithmic}
\usepackage{multirow}
\begin{document}
	
	\title{Delving into Crispness: Guided Label Refinement\\for Crisp Edge Detection}
	
	\author{
		Yunfan Ye,
		Renjiao Yi,
		Zhirui Gao,
		Zhiping Cai$^{*}$,
		Kai Xu$^{*}$\thanks{* Joint Corresponding authors.} \\
		National University of Defense Technology\\
		
		
			\thanks{This paper was produced by the IEEE Publication Technology Group. They are in Piscataway, NJ.}
			\thanks{Manuscript received July 27, 2022; revised April 17, 2023.}}
		
		\markboth{Journal of \LaTeX\ Class Files,~Vol.~14, No.~8, August~2021}%
		{Shell \MakeLowercase{\textit{et al.}}: A Sample Article Using IEEEtran.cls for IEEE Journals}
		
		
		\maketitle

		\begin{abstract}
			Learning-based edge detection usually suffers from predicting thick edges. Through extensive quantitative study with a new edge crispness measure, we find that noisy human-labeled edges are the main cause of thick predictions. Based on this observation, we advocate that more attention should be paid on label quality than on model design to achieve crisp edge detection. To this end, we propose an effective Canny-guided refinement of human-labeled edges whose result can be used to train crisp edge detectors. Essentially, it seeks for a subset of over-detected Canny edges that best align human labels. We show that several existing edge detectors can be turned into a crisp edge detector through training on our refined edge maps. Experiments demonstrate that deep models trained with refined edges achieve significant performance boost of crispness from $17.4\%$ to $30.6\%$. With the PiDiNet backbone, our method improves ODS and OIS by $12.2\%$ and $12.6\%$ on the Multicue dataset, respectively, without relying on non-maximal suppression. 
			We further conduct experiments and show the superiority of our crisp edge detection for optical flow estimation and image segmentation.
		\end{abstract}
		
		\section{Introduction}
		Edge detection is a longstanding vision task for detecting object boundaries and visually salient edges from images. As a fundamental problem, it benefits various downstream tasks ranging from object proposal \cite{zitnick2014edge, arbelaez2014multiscale}, optical flow \cite{revaud2015epicflow}, semantic segmentation~\cite{bertasius2016semantic, cheng2020boundary} and image inpainting \cite{xiong2019foreground, nazeri2019edgeconnect} among others. 
		
		Traditional methods extract edges based on local features such as gradients \cite{kittler1983accuracy, canny1986computational}. Recently, deep learning methods have achieved significant performance boost through learning more global and contextual features. Representatives include HED \cite{xie2015holistically}, RCF \cite{liu2017richer} and BDCN \cite{he2019bi}, which are based on the VGG \cite{simonyan2014very} architecture with multi-layer deep supervision.
		Although these methods attain good accuracy, they all suffer from the issue of predicting thick edges and thus heavily rely on the post-process of non-maximal suppression (NMS). On the other hand, NMS finds difficulty in handling thick edges spatially close to each other due to ambiguity.

		\begin{figure}[h]
			\centering
			\includegraphics[width=\linewidth]{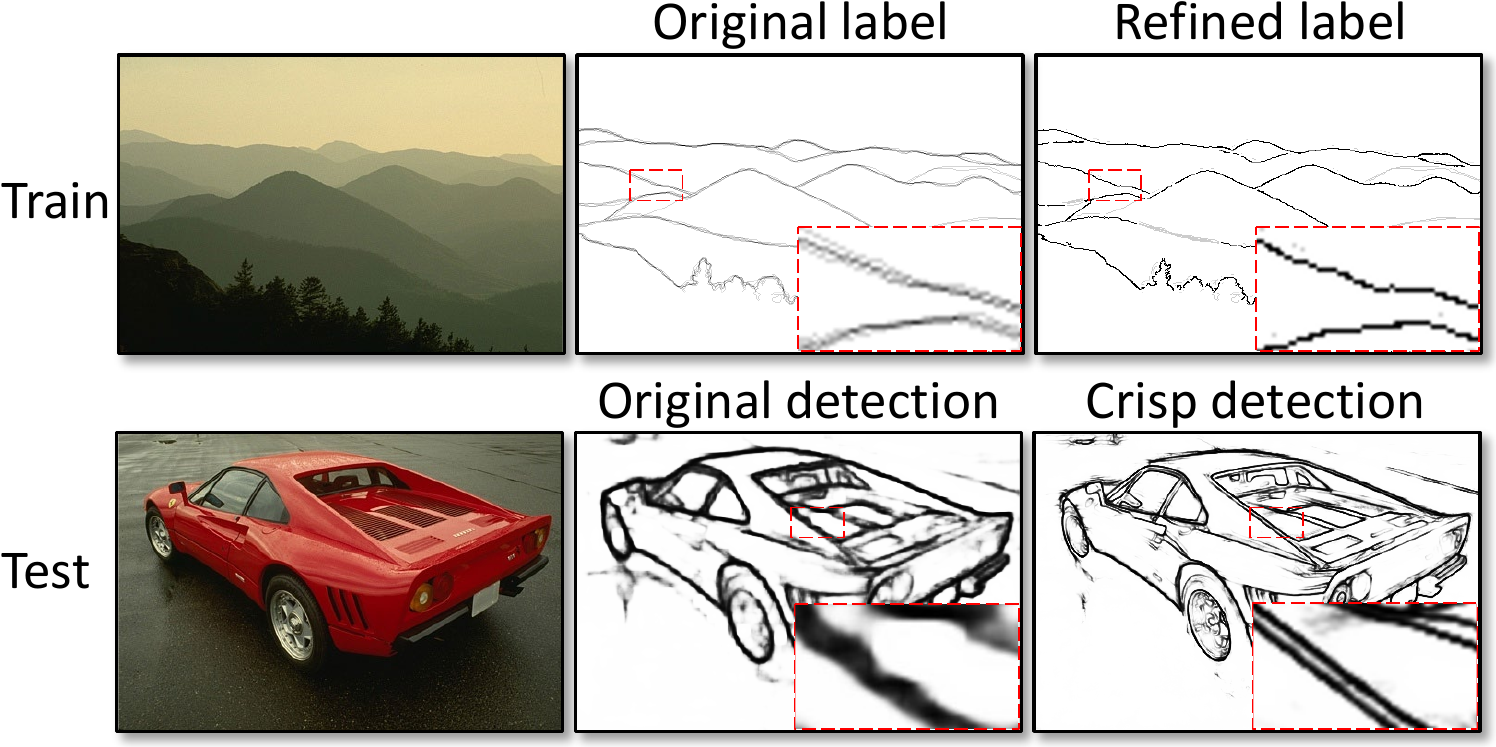}
			\caption{We propose a guided refinement strategy for human-labeled edge maps. The refined edge maps can be used to train crisp edge detection (bottom-right) without post-processing.}
			\label{fig:teaser}
		\end{figure}
		
		There have been some works studying how to detect \emph{crisp} edges. Wang et al. \cite{wang2017deep} propose a refinement architecture to progressively increase the resolution of feature maps to generate crisp edges. Deng et al. \cite{deng2018learning} believe the widely used weighted cross-entropy loss \cite{xie2015holistically} is the main cause of thick edges and propose the Dice loss to tackle the problem. Compared to cross-entropy loss, however, Dice loss lacks precise localization \cite{cheng2020boundary} and tends to yield weaker edge response \cite{deng2018learning}. SEAL \cite{yu2018simultaneous} and STEAL \cite{acuna2019devil} refine human labels based on active contours during training, leading to crisper edges in the task of semantic edge detection.
		
		In this paper, we observe through extensive empirical study that the noise of human labels is the main cause of thick edges by deep learning methods. It is very often that human labels deviate from the true edges due to imperfect human line drawing. We find that the more deviation the training labels have the thicker the predictions are made. Based on this observation, we advocate that more attention, if not all, should be paid on label quality rather than on model or loss design. Therefore, we propose an iterative Canny-guided refinement of human-labeled edges which can be used to learn crisp edge detection. We show that several existing edge detectors can be turned into a crisp edge detector through training on the refined edge maps, even for those using cross-entropy loss.
		Examples can be seen in Figure~\ref{fig:teaser}.

		To study edge crispness quantitatively, we propose the first crispness measure of edge detection based on the ratio of edge pixel values after and before NMS.
		Based on the measure, we conducted experiments on the BSDS500 dataset \cite{arbelaez2010contour} and observe the performance change of a detector trained on Canny labels perturbed with synthetic offsets of varying levels. The results verified that thick edges are mainly caused by noisy edge labels, and can be further aggravated when trained with cross-entropy loss.
		
		To realize effective edge refinement, for each training image, we first overlap the human-labeled edge maps with the corresponding over-detected Canny edge map and take the intersection of the two. This results in an edge map with edge segments coinciding with the true edges. We then correct the edge map by completing the segments with a pre-trained edge inpainting network. The inpainting network takes as input the gray-scale image of the input, an edge map to be completed, and a completion mask, and outputs a completed edge map. The mask is the difference between the current input edge map and the human labels, indicating which parts of the edge map need to be completed. Such completion essentially corrects the deviated portion of an edge through ``guessing''. The completed edges, after intersecting with Canny, are then fed into the inpainting network for the next-round refinement. The above process repeats until the change of the completion mask is negligible, implying that all human-labeled edges are collected and corrected. In essence, the interactive refinement seeks for a subset of the over-detected Canny edges aligning with human labels.
		
		In experiments, we apply our edge refinement on several public datasets (including BSDS500 \cite{arbelaez2010contour}, Multicue \cite{mely2016systematic} and BIPED  \cite{poma2020dense}) and train several edge detectors (e.g., PiDiNet \cite{su2021pixel}) on the refined edge maps. The results demonstrate that the models trained with refined edges achieve significant performance boost of crispness from $17.4\%$ to $30.6\%$ on the public datasets. Furthermore, it brings $12.2\%$ and $12.6\%$ improvement for ODS and OIS, respectively on the Multicue dataset without any post-processing, compared with the original model. 
		Besides the above experiments, we further applied our crisp edge detection to other vision tasks such as optical flow estimation and semantic segmentation. By integrating with edge-based algorithms Epicflow~\cite{revaud2015epicflow} and BNF~\cite{bertasius2016semantic}, we show that crisp edge detection is essential to those vision tasks and can boost their original performance.
		
		Our main contributions are as follows:
		\begin{itemize}
			\item Through quantitatively studying the impact of label noise on the crispness of edge detection, we observe that less label noise leads to crisper edges. In doing so, we propose the first crispness measure.
			\item We propose a novel Canny-guided iterative edge refinement method to correct human-labeled edges.
			\item We conduct extensive experiments of the proposed method and verify that corrected labels lead to significant improvement of edge crispness for existing detectors, and benefit other vision tasks.
		\end{itemize}

		\section{Related Work}
		\subsubsection{Edge detecion}
		The task of edge detection aims to extract object boundaries and visually salient edges from natural images. Traditional edge detectors focus on computing the image gradients to generate edges such as Sobel \cite{kittler1983accuracy} and Canny \cite{canny1986computational}. They often suffer from noisy pixels without semantic understanding. While learning-based methods start to train detectors to create semantic contours like \cite{arbelaez2010contour, dollar2014fast}. Recently, edge detection is booming with the success of Convolutional Neural Networks (CNNs). Based on VGG \cite{simonyan2014very} architecture, networks like \cite{xie2015holistically, liu2017richer, he2019bi} can achieve better precision performance with more semantic information. Efforts have also been made to design lightweight architectures
		for efficient edge detection including \cite{wibisono2020fined, poma2020dense, su2021pixel}. 
		
		While both precision performance and inference efficiency are improved, the issue of thick edges for result edge maps remains a problem. To evaluate the crispness of edges, \cite{wang2017deep} observe the degree of performance change by decreasing the maximal tolerant distance during the benchmark, and \cite{deng2018learning, huan2021unmixing} evaluate the edge maps for each model twice, one of the standard process and one of removing the post-processing. Such strategies to evaluate the edge crispness are all indirect and can not be measured by a certain value, which is inconvenient for comparison and study. To obtain crisper edges, previous works have made efforts on loss functions \cite{deng2018learning, acuna2019devil, huan2021unmixing}, refining labels during the training process \cite{yu2018simultaneous, acuna2019devil} and introducing extra modules \cite{wang2017deep, huan2021unmixing}. Though they can effectively obtain crisper edges, there are still some drawbacks including the balance of extra loss functions, timing decisions during the training process, or the introduction of extra modules and costs. In this paper, we explore a simple yet effective method for pre-processing the noisy labels to generate crisp edges concisely and propose the first measurable crispness evaluation metric.
		
		\subsubsection{Image inpainting}
		The task of image inpainting aims to fill missing regions in images with existing content. Driven by the progress of CNNs, the first deep learning method \cite{pathak2016context} adopts an encoder-decoder architecture and is improved in \cite{iizuka2017globally} by introducing a series of dilated convolution layers \cite{yu2015multi}. Some two-stage works attempt to improve the final inpainting results by introducing extra information like edges in the missing regions \cite{xiong2019foreground, nazeri2019edgeconnect, ren2019structureflow}. Downstream works including 3D Photography \cite{shih20203d} can also benefit from inpainted edges. Inspired by those works, we also adopt the edge inpainting network in our proposed edge refinement method. Different from previous works, we apply the inpainting locally and adaptively based on the specific human labels guided by Canny.
		
		\subsubsection{Edges for vision tasks}
		As a fundamental vision task, edge detection can benefit kinds of mid-level and higher-level vision tasks. There are commonly two schemes to leverage edge detection, directly adopting edge maps as one of the inputs, or making edge detection play as an auxiliary task during the learning process to impose edge-preserving consistency. 
		
		For the first scheme, edge maps and the specific vision task are decoupled. For example, Edge-Boxes~\cite{zitnick2014edge} and MCG~\cite{arbelaez2014multiscale} generate object bounding box proposals using edges, which can provide a sparse yet informative representation to indicate objects' positions. 
		EpicFlow~\cite{revaud2015epicflow} targets at large displacements and computes optical flow by leveraging edges to interpolate matches of adjacent video frames, based on the observation that motion discontinuities appear most of the time at image edges. 
		BNF~\cite{bertasius2016semantic} introduces boundary information to enhance semantic segment coherence and improve object localization based on the initial semantic segmentation.
		
		For the second scheme, edge detection serves as an auxiliary task and is integrated with the specific vision task. Image inpainting~\cite{xiong2019foreground, nazeri2019edgeconnect} leverage hallucinated edges of the missing region to avoid over-smoothed blurry and can reproduce filled regions of edge-preserving details. Moreover, by integrating specifically designed edge detection branches, vision tasks including depth completion~\cite{huang2019indoor}, depth estimation~\cite{zhu2020edge}, stereo matching~\cite{song2020edgestereo}, semantic segmentation~\cite{cheng2020boundary}, instance segmentation~\cite{kirillov2017instancecut} and glass-like object segmentation~\cite{he2021enhanced} can all benefit from extra edge-preserving consistency. In this paper, we adopt EpicFlow~\cite{revaud2015epicflow} and BNF~\cite{bertasius2016semantic} to show the superiority of crisp edge detection.

		\section{Study of Crisp Edge Mechanisms}
		In order to find the cause of thick edges, we do a comprehensive study of crisp edge mechanisms. We first review the loss functions of current methods, as well as some crisp label refinement works. Our assumption is that misalignment in human annotations is the main reason for thick edges in deep-learning methods, instead of the weighted binary cross entropy (W-BCE) loss as some prior works~\cite{deng2018learning, cheng2020boundary, huan2021unmixing} claim. To verify our assumption. We conduct experiments on different levels of misalignment simulating human-labeled edge maps, with qualitative and quantitative comparisons. Furthermore, due to the lack of an evaluation metric for edge crispness, we also propose the first measurable crispness evaluation metric, and prove its effectiveness with several experiments. 
		

		\subsection{Brief Reviews of Crisp Edges}
		\subsubsection{A Brief Review of Loss Functions}
		The task of edge detection is a pixel-wise binary classification problem. Since the numbers of edge and non-edge pixels are highly imbalanced, with the majority of pixels being non-edges. The conventional cross-entropy loss would degenerate in this case by predicting all pixels as non-edges. 
		HED \cite{xie2015holistically} applied balancing weights to normal binary cross entropy loss named W-BCE loss to solve this problem. Based on it, RCF \cite{liu2017richer} proposed an annotator-robust loss function. For the $i$th pixel in the $j$th edge map with value $p_{i}^{j}$, let $y_{i}$ be the ground truth edge probability of $i$th pixel, the annotator-robust loss is calculated as: 
		\begin{equation}\label{eq:wce_loss}
			l_{i}^{j}=
			\left\{
			\begin{array}{lll}
				\alpha \cdot \log \left(1-p_{i}^{j}\right), &  if \ y_{i}=0,\\
				0, & if \ 0 < y_{i} < \eta, \\
				\beta \cdot \log p_{i}^{j}, & otherwise,
			\end{array}
			\right.
		\end{equation} 
		in which
		\begin{equation}
			\begin{array}{l}
				\alpha=\lambda \cdot \frac{\left|Y^{+}\right|}{\left|Y^{+}\right|+\left|Y^{-}\right|}, \\
				\beta=\frac{\left|Y^{-}\right|}{|Y^{+}|+|Y^{-}|},
			\end{array}
		\end{equation}
		where $Y^+$ and $Y^-$ denote the number of edge and non-edge pixels in the ground truth edge maps. $\lambda$ is a parameter to balance positive and negative samples. $\eta$ is a threshold to filter out the less confident edge pixels in ground truths, to avoid confusing the network. For an edge map, the final loss is $\mathcal{L}=\sum_{i}^{j} l_i^j$.
		
		Previous works often hold a common conjecture that, although the W-BCE loss makes the network trainable, it also causes the issue of thick edges by predicting too many edge pixels~\cite{wang2017deep, deng2018learning, cheng2020boundary}. Dice loss is proposed in \cite{deng2018learning}, which measures the overlap between predictions and ground truths. It is insensitive to the number of edge/non-edge pixels, alleviating the class-imbalance problem and helping the neural network predict crisper edges. However, Dice loss lacks precise localization \cite{cheng2020boundary} and tends to yield weaker edge response compared with W-BCE loss. Currently, there is no perfect solution in terms of loss functions for crisp edge detection. 
		
		\begin{figure*}[h]
			\centering
			\includegraphics[width=\linewidth]{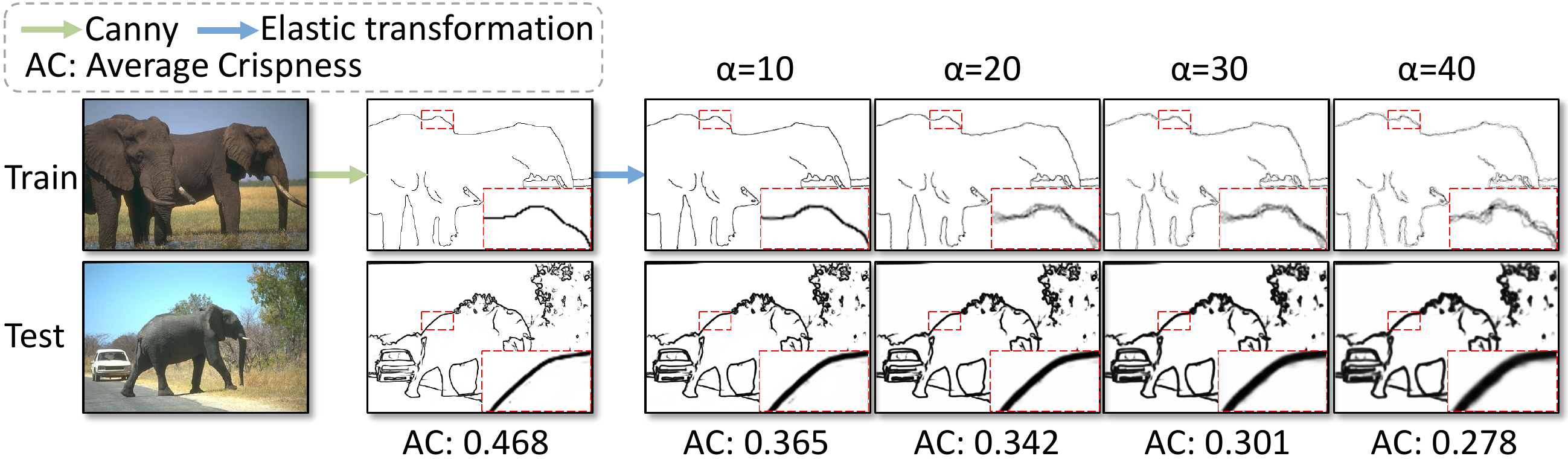}
			\caption{Impact of noisy labels. Zoom-in is recommended for better comparisons.}
			\label{fig:impact_of_noisy_labels}
		\end{figure*}

		\subsubsection{Brief Reviews of Label Alignment}
		Another assumption of predicting thick edges is the misalignment in edge maps annotated by different people. With inconsistent ground truth edge maps for the same data, the networks are confused and predict thick edges covering most annotated edge pixels. Therefore another direction to produce crisp edges is to refine the edge labels by removing misalignment. \cite{yang2016object} proposes to use dense conditional random fields (dense-CRF) to refine object contour in pre-processing. Its motivation is to improve the segmentation labels, and works not well for edge detection.  
		SEAL \cite{yu2018simultaneous} optimizes labels by formulating a computationally expensive bipartite graph min-cost assignment problem, while STEAL \cite{acuna2019devil} infers true object boundaries via a level set formulation which preserves connectivity and proximity. Both of them refine misaligned semantic contours during training and can make the contours of segmentation predictions crisper. However, such active alignment strategies heavily rely on the quality of network predictions, and have to decide the timing of label refinement during the training process. 
		
		Training with better-annotated labels can indeed make predicted edges crisper. The phenomenon indicates that the cross-entropy loss may not be the main reason leading to thick edges. Noisy labels, which are unavoidable in human annotations, are actually the main factor preventing crisp edges, the W-BCE loss just aggravates the situation. Several experiments are conducted to verify the assumption in the following sections. 
		
		\subsection{Crispness Evaluation}
		Although learning-based edge detection has made significant progress in precision, the quality of predicted edge maps is not satisfying without post-processing. Previous edge detection methods often focus on the ``correctness'' (the precision of edge pixel localization) instead of the ``crispness'' (the width of result edges). In the post-processing, a common Non-maximum Suppression (NMS) step is applied for all methods to ensure thin edges before evaluating on ground truth. However, NMS may introduce new problems such as connecting nearby edges. Predicting crisp edges directly by networks would be the best solution. 
		
		Before experiments, we define a crispness metric to measure how crisp the edge map is. There are several works trying to evaluate the crispness of predicted edge maps before, where \cite{wang2017deep} captures the crispness by decreasing the maximal permissible distance when matching ground-truth edges during the benchmark. \cite{deng2018learning, huan2021unmixing} examine the crispness by evaluating the edge maps before and after NMS, since predictions of high-crispness tend to perform better without the aid of NMS. However, both of them evaluate the crispness by observing the performance changes after some operations, which is indirect and can not be measured by a crispness value. In this section, to better explore crisp edge mechanisms, we propose a conceptually simple yet effective evaluation metric, termed as $Crispness$. The crispness for each edge map is calculated as the ratio of the sum of pixel values after NMS, to the sum of pixel values before NMS:
		\begin{equation}
			Crispness=\frac{\sum_{i=1}^{w \times h} NMS\left (E \right)}{\sum_{i=1}^{w \times h} E},
		\end{equation}
		where $E$ means the predicted edge map of size $w\times h$ and NMS represents the standard non-maximum suppression for edge detection. We take the average crispness (AC) when evaluating on a benchmark dataset. 
		
		The proposed crispness metric has several characteristics. First, it is lightweight and time-saving with about 100 FPS when evaluating on the BSDS500 dataset \cite{arbelaez2010contour}. Second, a crispness value is calculated and ranges from 0 to 1, for more straightforward comparisons. For an edge map where the edges are at a single-pixel width, the crispness would be close or equal to 1. The crisper the edge is, the larger the crispness would be.

		\subsection{Impact of Noisy Labels}\label{sec: noisy_labels}
		To verify the assumption of noisy labels as the main reason for thick edges, 
		we conduct experiments to measure the impact of noisy labels (label misalignment) on final crispness.
		
		We apply the Canny \cite{canny1986computational} detector to images to simulate real ground truth with no offsets, since the edge pixels produced by Canny are all single-pixel width and precisely lie in the positions where gradients change most sharply. 
		Then, we transform the Canny edge maps by moving pixels locally around using random displacement fields, called elastic transformation~\cite{simard2003best}, which is widely used for data augmentation in computer vision tasks~\cite{jackson2019style, zhao2020differentiable}. The parameter $\alpha$ controls the strength of the displacement, higher values mean that pixels are moved further. We perform elastic transformation for each $\alpha$ several times, and take their average to simulate human annotations from different people. 
		
		PiDiNet \cite{su2021pixel} is adopted as the edge detection backbone in the experiment. 
		We choose BSDS500 dataset \cite{arbelaez2010contour} to explore the impact of noisy labels. To be specific, we train PiDiNet on the training set of BSDS500 with labels all generated by Canny, and its different elastic transformation versions as comparing groups, then evaluate the average crispness of the different network versions on the testing set. We use the commonly-used W-BCE loss in Equation \ref{eq:wce_loss} to train networks. The results are demonstrated in Figure \ref{fig:impact_of_noisy_labels}. We can see that the smaller the $\alpha$ is, the crisper and better the result edges are, both qualitatively and quantitatively. It means larger annotation offsets (misalignment) cause thicker edges, and the crispness is close to the one training from human annotations when $\alpha$=40. 
		The edge map predictions by the network trained from Canny labels are much crisper than other groups, which means networks trained with W-BCE loss can still get crisp edges. The label misalignment, instead of the loss function, is the real reason for thick edges.  
		\begin{figure}[h]
			\centering
			\includegraphics[width=\linewidth]{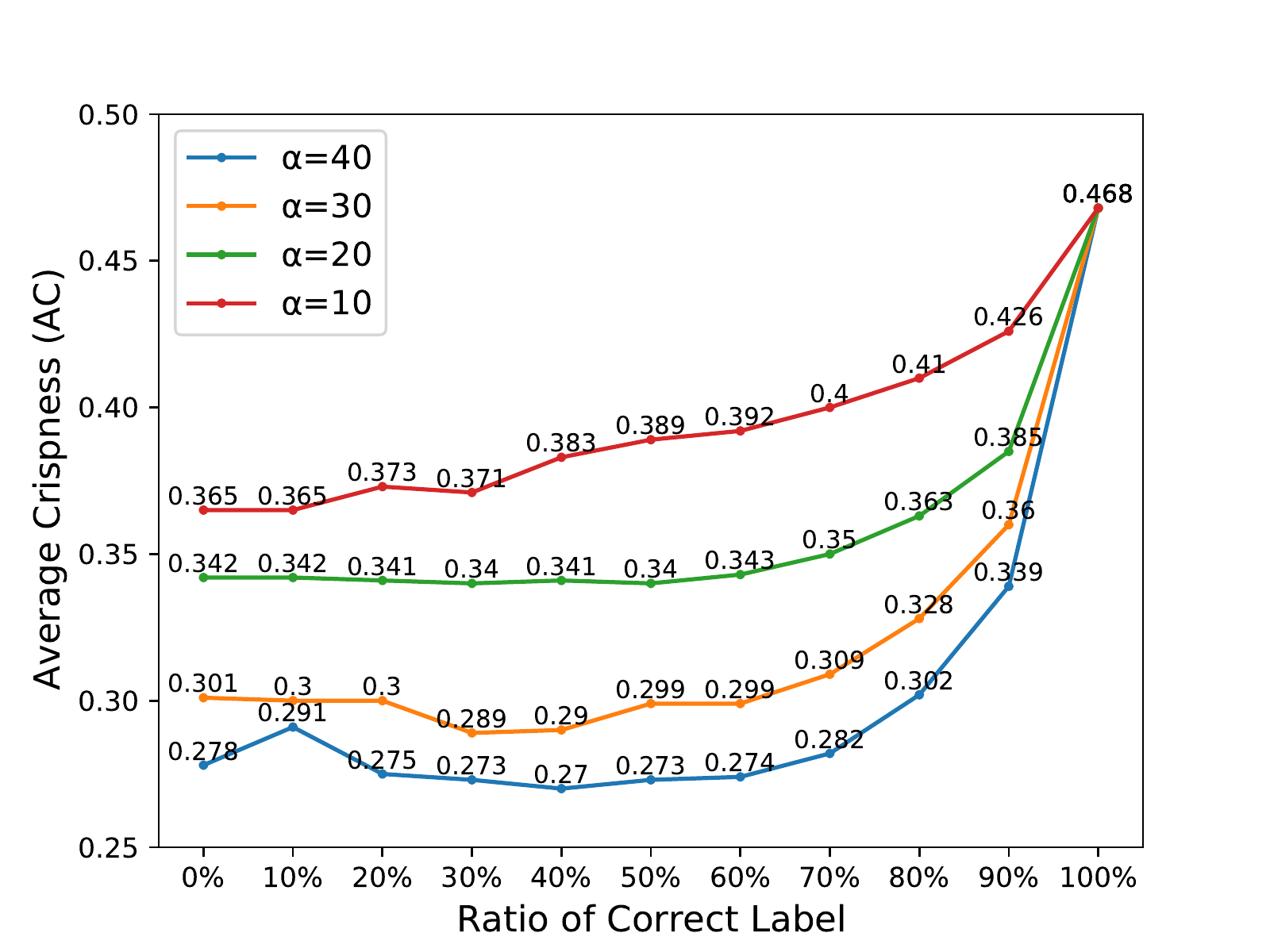}
			\caption{Average Crispness for different percentage of correct training labels.}
			\label{fig:impact_of_noisy_labels_line_chart}
		\end{figure}
		
		\begin{figure*}[h]
			\centering
			\includegraphics[width=\linewidth]{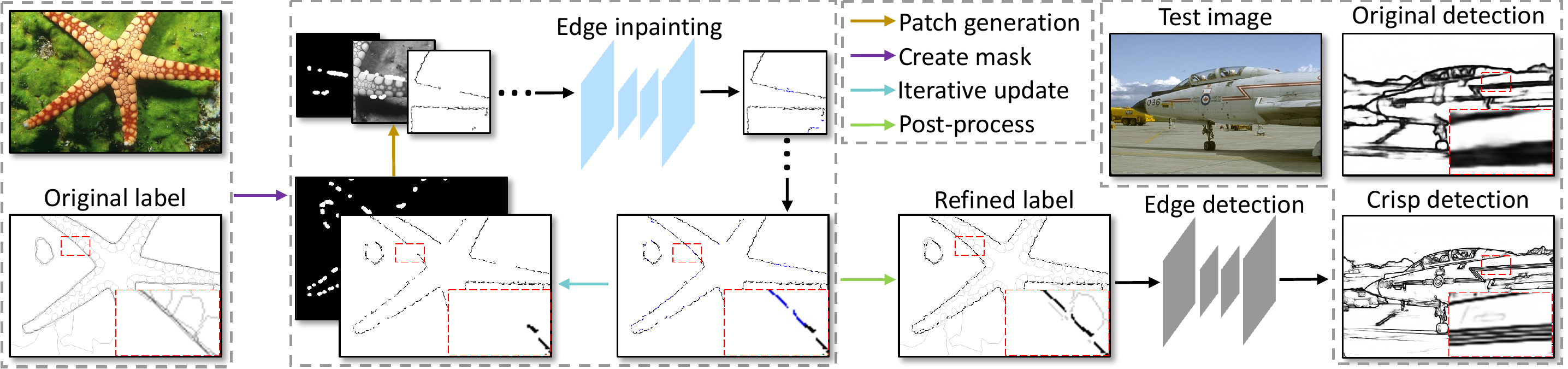}
			\caption{The pipeline of our iterative edge label refinement and crisp edge detection.}
			\label{fig:patch_based_edge_refine_architecture}
		\end{figure*}
		
		In real scenarios, it is common that only part of human annotations are noisy while some others are carefully annotated. Thus, we further conduct experiments with different percentages of elastic-transformed Canny labels with a fixed $\alpha$. The original Canny results are treated as ``correct'' labels. As shown in Figure \ref{fig:impact_of_noisy_labels_line_chart}, except for the line $\alpha=10$, when we gradually increase the percentage of correct labels, the AC changes slowly before reaching 80\%, while the AC increases dramatically between 90\% and 100\%. 
		For the line $\alpha=10$, which means the label offsets are very small and most labels can be considered correct enough, the AC gradually increases in pace with the ratio of correct labels. Such interesting phenomenons indicate that the result edges can be crisp enough only when most of the training labels ($>$90\%) have no offsets. It explains why most previous works consider W-BCE loss to be the main cause of thick edges, since it is hard or even impossible for human annotations to have ``correct'' labels of more than 90\% in edge detection. We come to the conclusion that the issue of thick edges generated by neural networks is mainly caused by noisy human annotations, and aggravated by the W-BCE loss. With the premise of such a conclusion, we try to propose a general and simple edge label refinement method as a pre-processing before training edge detection networks. With less-noisy labels, the performance of the trained network would be much better in both precision and crispness. Moreover, since changes only happen in original labels, our method can be easily applied to any edge detection backbone, leading to a boost of crispness. 
		
		\section{Crisp Edge Refinement Pipeline}
		
		\subsection{Patch-based Edge Refinement}
		We aim to propose a technically simple but effective method that can directly pre-process and refine noisy human labels without introducing any extra modules and complicated loss combinations, enabling other backbone networks to generate crisper edges by training with refined labels. The label refinement pipeline is easy to use and can be integrated into any edge detection network, with minimal changes to the original labels. Based on the premise that Canny edges lie in the exact positions where gradients change most sharply, bringing no ambiguity when training with W-BCE loss, unlike human-annotated edges, we naturally develop the label refinement method facilitated by Canny edges.  
		Specifically, we apply the Canny detector of a very low threshold to the multiple blurred images of the input image with different blur strategies, and fuse their results as the final Canny edges. Edges in the obtained Canny edge map are over-detected, which means the real crisp ground truth edges can be considered as a subset of these edges. Our task is to iteratively refine this subset of edges as completely as possible without any offsets, and get the crisp ground truth edge map at last. 
		
		To do so, we first calculate the overlap (Hadamard product) of human labels annotated by different people and the over-detected Canny edge map as the initial edge map. Apparently, most human-annotated edge maps have offsets to the corresponding Canny edges, and cannot match Canny edge pixels precisely. As a result, their overlap would have many discontinuities. One straightforward approach is to inpaint the discontinuous edges through adversarial networks. Specifically, we build our edge inpainting model upon image inpainting methods in \cite{liu2018image, nazeri2019edgeconnect, xiong2019foreground}. The key difference between our model and those recent two-stage approaches is that, their inpainting masks and the available contexts are static, while our method inpainting edges locally around each edge discontinuity adaptively. In other words, our inpaint masks are dynamic during the iterative label refinement pipeline.

\begin{figure}[h]
			\centering
			\includegraphics[width=\linewidth]{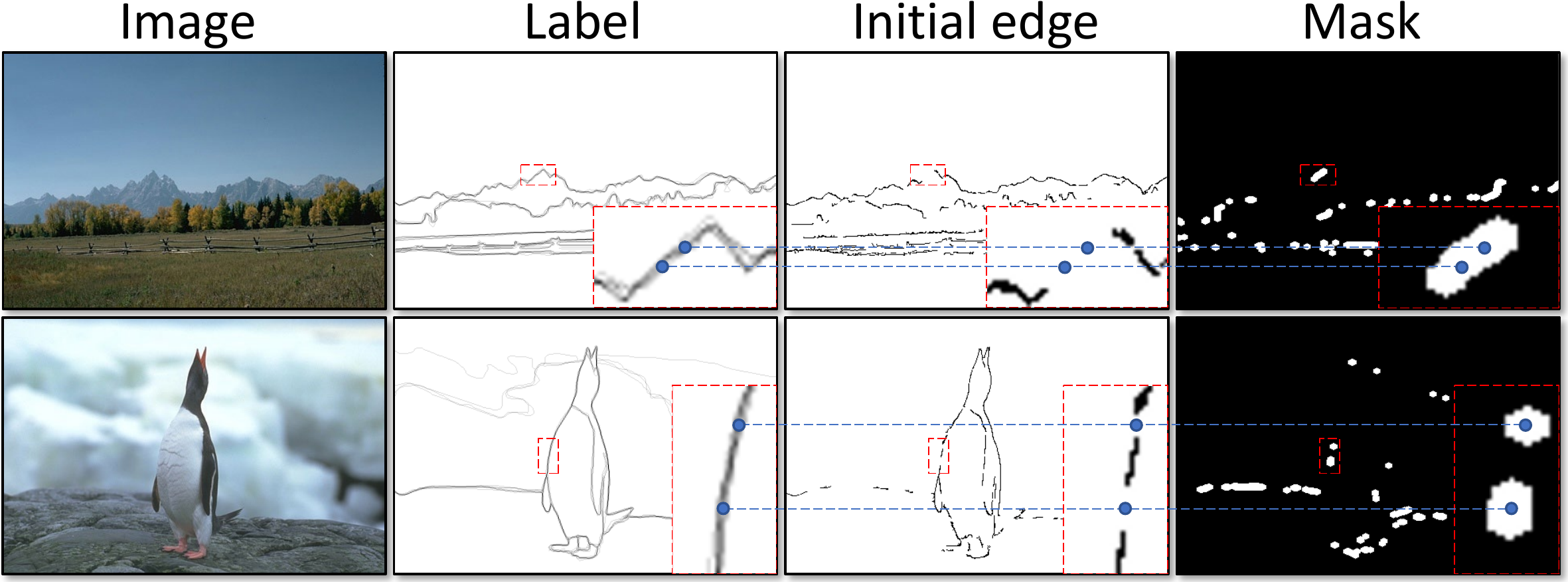}
			\caption{Examples for generating the inpainting masks for initial edges. We check each position of edge pixels in the human label, and treat this position as an inpainting pixel (blue points are examples of some inpainting pixels) if there are no edges around the position on the initial (refined) edge map. The inpainting mask is further generated by dilating all inpainting pixels.}
			\label{fig:mask_generation}
		\end{figure}

		We propose an adaptive patch-based edge refinement strategy to iteratively inpaint the missing edges from initial edge maps and compute the overlap with Canny edges, as illustrated in Figure \ref{fig:patch_based_edge_refine_architecture}. We first compute the Hadamard product of over-detected Canny edges from the blurred input image and human labels as initial edge, where there would be several discontinuous edges waiting for refinement (see Section~\ref{sec: ablation_dropout} for more ablations of robustness). In order to decide which pixels to inpaint on the inital (refined) edge map, we check the neighborhood of each edge pixel in human labels, if there are no edges nearby on the inital (refined) edge map, we consider the current position as an inpainting pixel. At last, we dilate the binary mask of inpainting pixels as the inpainting mask, examples are provided in Figure~\ref{fig:mask_generation}.  
		
		\begin{figure}[h]
			\centering
			\includegraphics[width=\linewidth]{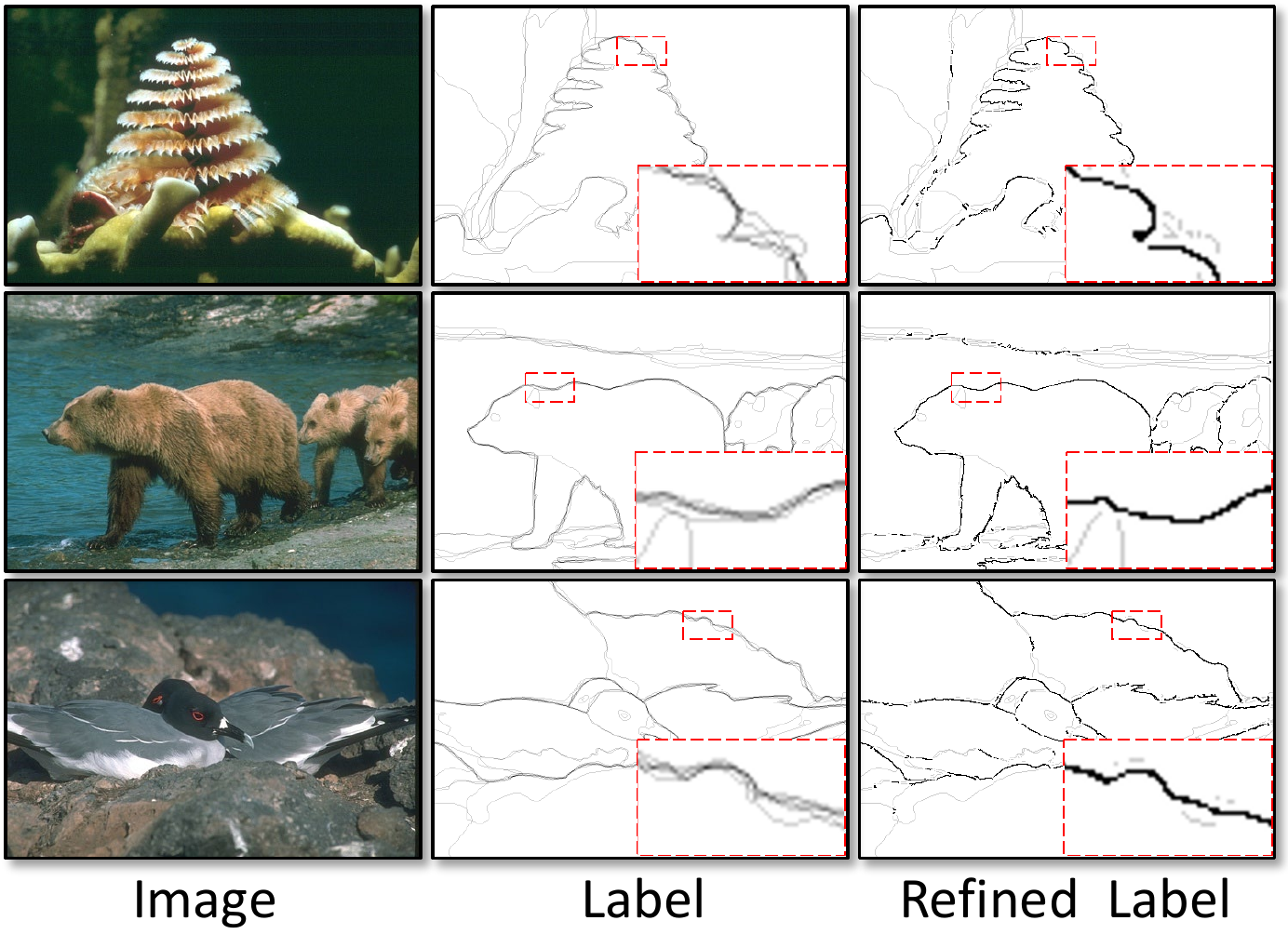}
			\caption{Examples from BSDS500 dataset about the raw human annotations and our refined labels.}
			\label{fig:raw_and_refine_edge}
		\end{figure}
		
		In the edge inpaint module, we adaptively inpaint the edge map by inputting the gray-scale image, the edge map of current iteration and the inpainting mask. The edge inpainting network is applied for each image patch and the inpainting process lasts for several iterations until no new edge is created in the most recent iteration. 
		
		In this work, we employ the number of connected areas in the inpainting mask as the indicator for termination.
		Sometimes during the refinement process, one big connected area in the inpainting mask could be occasionally divided into two smaller ones if only the center part has been refined. However, such extreme cases are rare even in each image. Since the refinement method only inpaints edge pixels inside the inpainting mask, it can be guaranteed that the inpainting mask is always non-increasing by nature. Therefore, as the iterative refinement proceeds, the overall number of connected areas in the inpainting mask will gradually decrease until convergence. 
		To avoid any potential infinite loop, there is also a maximum number of allowed iterations ($I_{max}$ in Algorithm~\ref{alg:edge_refinement}).
		
		At last, we set the values of those human annotated pixels that are not in the refined label to be smaller than $\eta$ (in Equation~\ref{eq:wce_loss}), so that the subsequent training ignores these unconfident pixels. 
		See Algorithm \ref{alg:edge_refinement} for detailed pipeline and formulations. Qualitative comparisons between original human annotations and our refined labels are illustrated in Figure \ref{fig:raw_and_refine_edge}.

		\begin{algorithm}[htb]
			\renewcommand{\algorithmicrequire}{\textbf{Input:}}
			\renewcommand{\algorithmicensure}{\textbf{Output:}}
			
			\caption{Iterative Edge Label Refinement.}
			\label{alg:edge_refinement}
			\begin{algorithmic}[1] 
				\REQUIRE ~~\\ 
				The original image $X$; The original edge label $Y$; The patch size $S$; The threshold $\eta$ as in Equation \ref{eq:wce_loss}; The maximum iteration number $I_{max}$. 
				\ENSURE ~~\\ 
				The well-refined new edge label $Y^{\prime}$. 
				
				\STATE Generate the initial edge $E$ by multiplying the over-detected Canny  edge map $C$ and the original label filtered by $\eta$:
				
				$C = Canny (Blur (X), thres\_low)$, 
				
				$Y[Y < \eta]]=0$,
				
				$E = C \odot Y$;

				\STATE Calculate the mask area $M$ where edge discontinuities need to be inpainted in $E_i$:
				
				$M = Create\_mask(Y, E)$;
				
				\FOR{round $i$ in $\left \{ 1,...,I_{max} \right \}$ }
				\STATE Generate patches ${P}=[{p_{1}}, {p_{2}},...,{p_{n}}]$ of patch size $S$ with the same center of each connected area in the mask $M$:
				$P = Create\_patches(M, S)$ ;	
				
				\FOR{each patch $p_n$ in $P$ }
				
				\STATE Inpainting the edge discontinuities by edge patch $E(p_n)$ and the grayscale of original image patch $X_{gray}(p_n)$ guided by mask patch $M(p_n)$. Then update the patch guided by $C$:
				
				$E^{\prime}(p_n) = Edge\_inpainting (E(p_n), X_{gray}(p_n), M(p_n))$,
				
				$E(p_n) = C \odot E^{\prime}(p_n)$;
				\ENDFOR 
				
				\STATE Terminate the iterative refinement process when the current number of connected areas $N_{connect}^{n}$ of mask $M$ is no less than it of last iteration:
				
				\textbf{if} $N_{connect}^{n} >= N_{connect}^{n-1}$ \textbf{then}
				
				\ \ \ \ \textbf{break};
				
				\textbf{end if}

				\ENDFOR 
				
				\STATE Post-processing to set a value smaller than $\eta$ in positions where any positive pixels in $Y$ that is negative in $E$:
				
				$Y^{\prime} = Post\_process(Y, E, \eta)$.
			\end{algorithmic}
		\end{algorithm}
		
		\subsection{Upscaling Strategy for Crisp Edge Detection}\label{sec:upscale}
\begin{figure}[h]
			\centering
			\includegraphics[width=\linewidth]{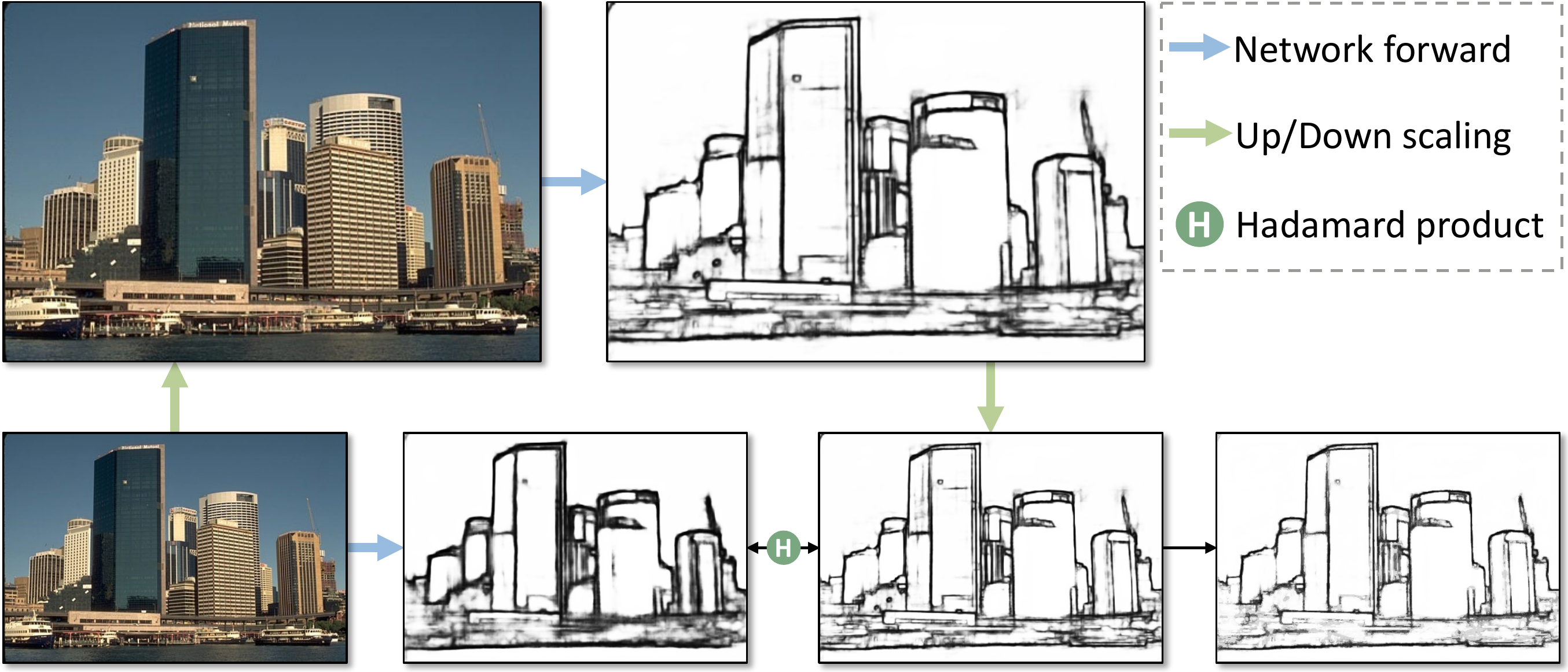}
			\caption{The pipeline of crisp edge algorithm with the upscaling strategy.}
			\label{fig:upscale_crisp}
		\end{figure}

		Inspired by the multi-scale strategy in \cite{liu2017richer} which averages the resulting edge maps of input image pyramids to improve the performance, we propose an upscaling strategy to further make edges crisper. The original image and its up-scaled version are fed into the backbone edge detection network to get corresponding edge maps. Then, we downscale the up-scaled resulting edge map and restore it to its original size. A Hadamard multiplication is applied to the two edge maps to get a crisper result. This process is illustrated in Figure \ref{fig:upscale_crisp}. With a fixed receptive field size, there is relatively less available context on the up-scaled input image compared to the image at its original size. Therefore the network is prone to generate more detailed edges. For images at the original or smaller sizes, there are more contextual information and the networks tend to create more meaningful and contextual but thicker boundaries. Through the upscaling strategy, we take the advantage of different input sizes by fusing the edge maps. 
		
		This strategy has many merits. 
		First, compared to the standard NMS, the upscaling strategy is end-to-end. While NMS has trouble separating adjacent edges that are glued together in the raw predictions, our method can clearly generate two separated edges. The upscaling strategy is a quick and effective way to refine thick edge predictions of edge detection networks to be crisper. 
		This strategy introduces more computational and time costs to the whole pipeline, thus considering both the accuracy and speed, we set the upscaling factor as 1.5 for all cases. See Section \ref{sec:Experiments} for more details.
		
		\section{Experiments} \label{sec:Experiments}
		\subsection{Datasets and Implementation Details}\label{sec:dataset}
		\subsubsection{Benchmark Datasets}
		We evaluate the proposed crisp edge method on three widely used datasets,
		BSDS500 \cite{arbelaez2010contour}, Multicue \cite{mely2016systematic}, and BIPED \cite{poma2020dense}. BSDS500 consists of three splits of 200, 100 and 200 images for training, validation and testing, respectively. Each image is labeled by 4 to 9 annotators and the final edge ground truth is computed by taking the average of them. We adopt the training and validation sets of 300 images for training with data augmentation of flipping (2×), and rotation (4×). Multicue consists of images from 100 challenging natural scenes. Each scene contains a sequence of images from the left and right views, captured by a stereo camera. Only the last frame of every left-view sequence is labeled by annotators. Each of these 100 images is annotated by several people as well. 
		We randomly split the 100 images into training and evaluation sets, consisting of 80 and 20 images respectively. The same as BSDS500 dataset, the average edge map of several annotations is computed as the ground truth. 
		Images in the Multicue dataset are at higher resolutions, which are 1280 $\times$ 720. Thus we augment each image by flipping (2×), cropping (3×), and rotation (8×), leading to a training set that is 48 times larger than the original dataset. BIPED contains 250 annotated images of outdoor scenes, splitting into a training set of 200 images and a testing set of 50 images. All images are carefully annotated by experts in the computer vision field to get crisp edge labels at single-pixel width. The resolution of BIPED is also 1280 $\times$ 720, and images are augmented with flipping (2×), cropping (3×), and rotation (8×).  
		
		\subsubsection{Backbone networks}
		Our crisp edge detection pipeline can be integrated with any edge detection backbones, here we adopt two most recent edge detectors, PiDiNet~\cite{su2021pixel} and DexiNed~\cite{poma2020dense}, to validate the effectiveness of the proposed method. 
		For the edge inpainting network, we adopt an existing architecture\cite{nazeri2019edgeconnect} to fill the edges in missing regions. The network takes the grayscale image, the incomplete edge map and the inpainting mask as inputs, and outputs an edge map filled with predicted edges in missing areas. 
		
		\subsubsection{Evaluation Metrics}
		For general edge detection, the predicted edge maps are in gray-scale with pixel values in [0, 1], while the ground truth are binary in nature with each pixel labeled as either 0 (non-edge) or 1 (edge).
		To evaluate the precision, recall, and F-score, an optimal threshold that binarizes the predicted edge map is needed. 
		
		Following prior works, we compute the F-scores of Optimal Dataset Scale (ODS) and Optimal Image Scale (OIS), which are two strategies of determining the optimal thresholds to binarize predicted edge maps. ODS employs a fixed threshold for all images in the dataset while OIS chooses an optimal threshold for each image. F-scores are computed by $F = \frac{2\cdot P\cdot R}{P+R}$, where $P$ denotes precision and $R$ denotes recall. 
		Considering both accuracy and time cost, $P$ and $R$ are calculated for each threshold from 0.01 to 0.99 for every image. If multiple ground truth labels are given, such as the BSDS500 \cite{arbelaez2010contour} and Multicue \cite{mely2016systematic}, $P$ is the number of edges that can match all the labels, divided by the number of edges in the binarized prediction result; $R$ is the sum of the number of edges that can match each label, divided by the sum of the number of edges from each label.
		For ODS and OIS, the maximum allowed distances between corresponding pixels from predicted edges and ground truths are set to 0.0075 for all experiments. 
		
		ODS and OIS emphasize more on the precision of edge pixel localization, while the main focus of this paper is improving the crispness with comparable performance. It is the reason for introducing a new metric Crispness in this paper. Noises on human labels are unavoidable, and there is no way to get true edge labels without any noise. Therefore, we apply ODS and OIS as a reference for accuracy performance and also report the proposed crispness measure as the main metric. Crispness is averaged among all images in the test dataset (``AC'' in tables). The proposed method leads to significantly higher Crispness where ODS and OIS are comparable.

		\subsubsection{Implementation Details}
		Our method is implemented in the Pytorch~\cite{paszke2019pytorch} environment. While testing on PiDiNet and DexiNed backbones, we keep their original settings for a fair comparison. W-BCE loss in Equation \ref{eq:wce_loss} is adopted for training with Adam optimizer \cite{kingma2014adam}, at an initial learning rate of 0.005, decaying in a multi-step way. $\lambda$ is set to 1.1 and the threshold $\eta$ is set to 0.3 for all datasets while computing W-BCE losses.
		The architecture of the edge inpainting model is the same as~\cite{nazeri2019edgeconnect} and trained on the Place2 \cite{zhou2017places} dataset. All images are resized to 256 $\times$ 256 for training with a batch size of 8, optimized by Adam optimizer. Therefore, we set the patch size as 256 $\times$ 256 as well while applying the locally patch-based edge refinement.
		
		\begin{figure*}[h]
			\centering
			\includegraphics[width=\linewidth]{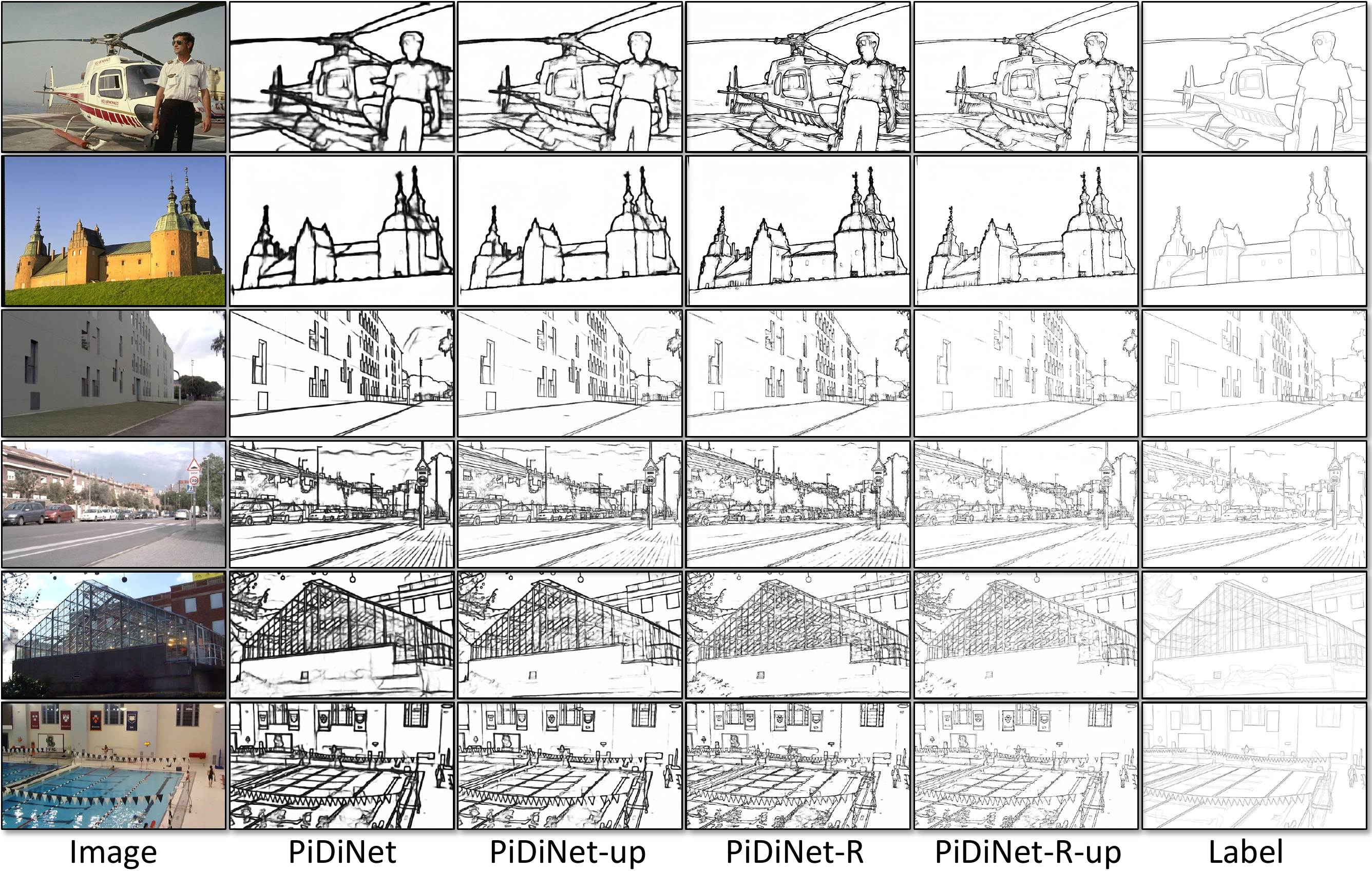}
			\caption{Qualitative results on datasets of BSDS500 (the first two rows), BIPED (the second two rows) and Multicue (the last two rows) before NMS. Edge maps become much crisper with the aid of our edge refinement method (``-R'') and upscaling strategy (``-up''). Zoom-in is recommended to see the crispness details.}
			\label{fig:comparisons_on_datasets}
		\end{figure*}
		
		\subsection{Evaluations}
		
		In this section, we report the statistical evaluations of our label refinement method compared to two backbone methods PiDiNet and DexiNed. Experiments are conducted on BSDS500, Multicue and BIPED datasests. The quantitative evaluations on three datasets are in Table~\ref{tab:comparison_on_datasets}. The analysis is organized in two aspects, after and before NMS, with and without the edge refinement method. Qualitative comparisons are presented in Figure~\ref{fig:comparisons_on_datasets}. 
		
		As expected, we (``-R'') achieve significant performance boosting on average crispness in all cases when training with our refined labels, both qualitatively and quantitatively. For the backbone of PiDiNet, our method improves the average crispness significantly by 17.4\%, 22\%  and 28\% on BSDS, Multicue and BIPED datasets. For the backbone of DexiNed, our method also improves the average crispness by 8\%, 11.8\%  and 13.8\% on BSDS, Multicue and BIPED datasets, respectively. 
		
		\begin{table}[htbp]
			\centering
			\scalebox{0.95}{
				\begin{tabular}{c|c|c|c|c|c|c}
					\hline
					\multirow{2}{*}{Dataset} & \multirow{2}{*}{Methods} & \multicolumn{2}{c|}{after NMS} & \multicolumn{2}{c|}{before NMS} & \multirow{2}{*}{AC} \\
					\cline{3-6}          &       & ODS   & OIS   & ODS   & OIS   &  \\
					\hline
					\multirow{4}{*}{BSDS 500} & PiDiNet & \textbf{0.780}  & \textbf{0.794} & 0.763 & 0.776 & 0.250 \\
					& PiDiNet-R & 0.771 & 0.782 & \textbf{0.771} & \textbf{0.780} & \textbf{0.424} \\
					
					\cline{2-7}
					
					& DexiNed & {0.722}  & {0.746} & 0.701 & 0.723 & 0.277 \\
					& DexiNed-R & 0.719 & 0.744 & {0.708} & {0.730} & {0.357} \\
					\hline
					\multicolumn{1}{c|}{\multirow{4}{*}{Multicue}} & PiDiNet & 0.874 & 0.878 & 0.764 & 0.778 & 0.204 \\
					& PiDiNet-R & \textbf{0.892} & \textbf{0.907} & \textbf{0.886} & \textbf{0.904} & \textbf{0.424} \\
					
					\cline{2-7}
					
					& DexiNed & 0.855 & 0.868  & 0.748 & 0.775  & 0.274 \\
					& DexiNed-R & {0.863}  & {0.877}  & {0.849} & {0.863}  & {0.392} \\
					\hline
					\multirow{4}{*}{BIPED} & PiDiNet & \textbf{0.868} & \textbf{0.876} & 0.842 & 0.852 & 0.232 \\
					& PiDiNet-R & 0.863 & 0.873 & \textbf{0.863} & \textbf{0.873} & \textbf{0.512} \\
					
					\cline{2-7}
					
					& DexiNed & 0.841 & 0.853 & 0.831  & 0.842 &  0.295\\
					& DexiNed-R & {0.858} & {0.868} & {0.855} & {0.868} & {0.433} \\
					\hline
			\end{tabular}}
			\caption{Quantitative results for PiDiNet when training with original labels and refined labels. ``-R'' means training with refined labels and the same for other tables and figures.}
			\label{tab:comparison_on_datasets}%
		\end{table}%
		
		When non-maximum suppression (NMS) is adopted on predicted results, 
		for models using our label refinement pipeline (``-R'' in tables), F-scores of ODS and OIS are boosted on Multicue and BIPED dataset in terms of DexiNed backbone, and F-scores of ODS and OIS are boosted on Multicue in terms of PiDiNet backbone.  
		We notice slight performance drops of around 1\% for some conditions. The reason could be that not all human annotations are correct enough including the test datasets, and our refinement method actually changes some edge pixels with relatively big offsets, which may be unfavorable to the quantitative evaluation of our method. 
		
		Meanwhile, although NMS is a commonly used post-processing to get crisp edges, it is not differentiable, which cannot be integrated into end-to-end training and testing. Furthermore, NMS is time-consuming. It runs at 100 fps on BSDS, whereas the inference of PiDiNet runs at 92 fps, which means adopting NMS doubles the running time, while our label refinement method does not lead to any extra time when testing. For high-resolution datasets such as BIPED, NMS takes even longer time and runs at only 25 fps. All time are reported on an RTX 2080 Ti GPU. 
		
		Without NMS (``before NMS'' in tables), the performance of models training with original labels drops significantly, since thick edges tend to have high recall but very low precision, leading to low F-scores. With our refinement method, the performance boosts significantly because crisp edges predicted by networks already have a better balance between precision and recall.

		\begin{table}[htbp]
			\centering
			\scalebox{0.95}{
				\begin{tabular}{c|c|c|c|c|c|c}
					\hline
					\multirow{2}{*}{Dataset} & \multirow{2}{*}{Methods} & \multicolumn{2}{c|}{after NMS} & \multicolumn{2}{c|}{before NMS} & \multirow{2}{*}{AC} \\
					\cline{3-6}          &       & ODS   & OIS   & ODS   & OIS   &  \\
					\hline
					\multirow{3}{*}{BSDS 500} &PiDiNet& \textbf{0.780}  & \textbf{0.794} & 0.763 & 0.776 & 0.250 \\& PiDiNet-up & \textbf{0.780}  & 0.793 & \textbf{0.773} & \textbf{0.787} & 0.323 \\
					& PiDiNet-R-up & 0.774 & 0.785 & 0.771 & 0.780  & \textbf{0.522} \\
					\hline
					\multicolumn{1}{c|}{\multirow{3}{*}{Multicue}} & PiDiNet & 0.874 & 0.878 & 0.764 & 0.778 & 0.204 \\& PiDiNet-up & \textbf{0.890}  & \textbf{0.900}   & \textbf{0.880}  & \textbf{0.890}  & 0.284 \\
					& PiDiNet-R-up & 0.874 & 0.893 & 0.863 & 0.887 & \textbf{0.555} \\
					\hline
					\multirow{3}{*}{BIPED} & PiDiNet & \textbf{0.868} & \textbf{0.876} & 0.842 & 0.852 & 0.232 \\& PiDiNet-up & 0.860  & 0.869 & 0.850  & 0.864 & 0.318 \\
					& PiDiNet-R-up & 0.862 & 0.872 & \textbf{0.860}  & \textbf{0.870}  & \textbf{0.624} \\
					\hline
			\end{tabular}}
			\caption{Quantitative analysis for PiDiNet when applying the upscaling strategy (``-up'') described in Section~\ref{sec:upscale}.}
			\label{tab:comparison_on_datasets_up}
		\end{table}

		For the upscaling strategy for crisp edge detection, as shown in Table~\ref{tab:comparison_on_datasets_up}, this simple strategy (``-up'' in the table) is effective in further improving the crispness. On BIPED dataset, applying the upscaling strategy to models trained with our refined labels achieve better performance on all metrics of ODS, OIS and AC, and combining with refined labels leads to a crispness improvement of 30.6\%. Although on Multicue and BSDS datasets, combining label refinement and upscaling strategy leads to a slight performance drop (less than 1\% when NMS is not adopted), it still provides a significant boost on crispness. Comparing Table~\ref{tab:comparison_on_datasets} and \ref{tab:comparison_on_datasets_up}, both label refinement and upscaling strategy work effectively for generating more accurate and crisper edge maps. As illustrated in Figure~\ref{fig:comparisons_on_datasets}, combining both strategies also leads to significantly better (and crisper) visual results.

		\subsection{Ablation Study}
		\subsubsection{Ablations for refinement strategies}
		To demonstrate the effectiveness of our iterative Canny-guided edge refinement pipeline, we adopt the PiDiNet backbone and conduct several ablation studies on the BSDS500 dataset. The data splitting and augmentation are described in Section~\ref{sec:dataset}. 
		
		The initial edge labels are computed by simply taking the overlap of the over-detected Canny edges and the original human labels. We find that by simply computing a Hadamard product with Canny edges for all training data, 
		the average crispness is satisfactory enough, but the quantitative performance drops sharply without further optimization. It means, refining human labels by simply taking the overlap with Canny edges is not enough. After refining the initial edge once or iteratively by our pipeline, the predicted edges become both crisper and more accurate, which shows the effectiveness of each module in our designs. 
		
\begin{figure}[h]
			\centering
			\includegraphics[width=\linewidth]{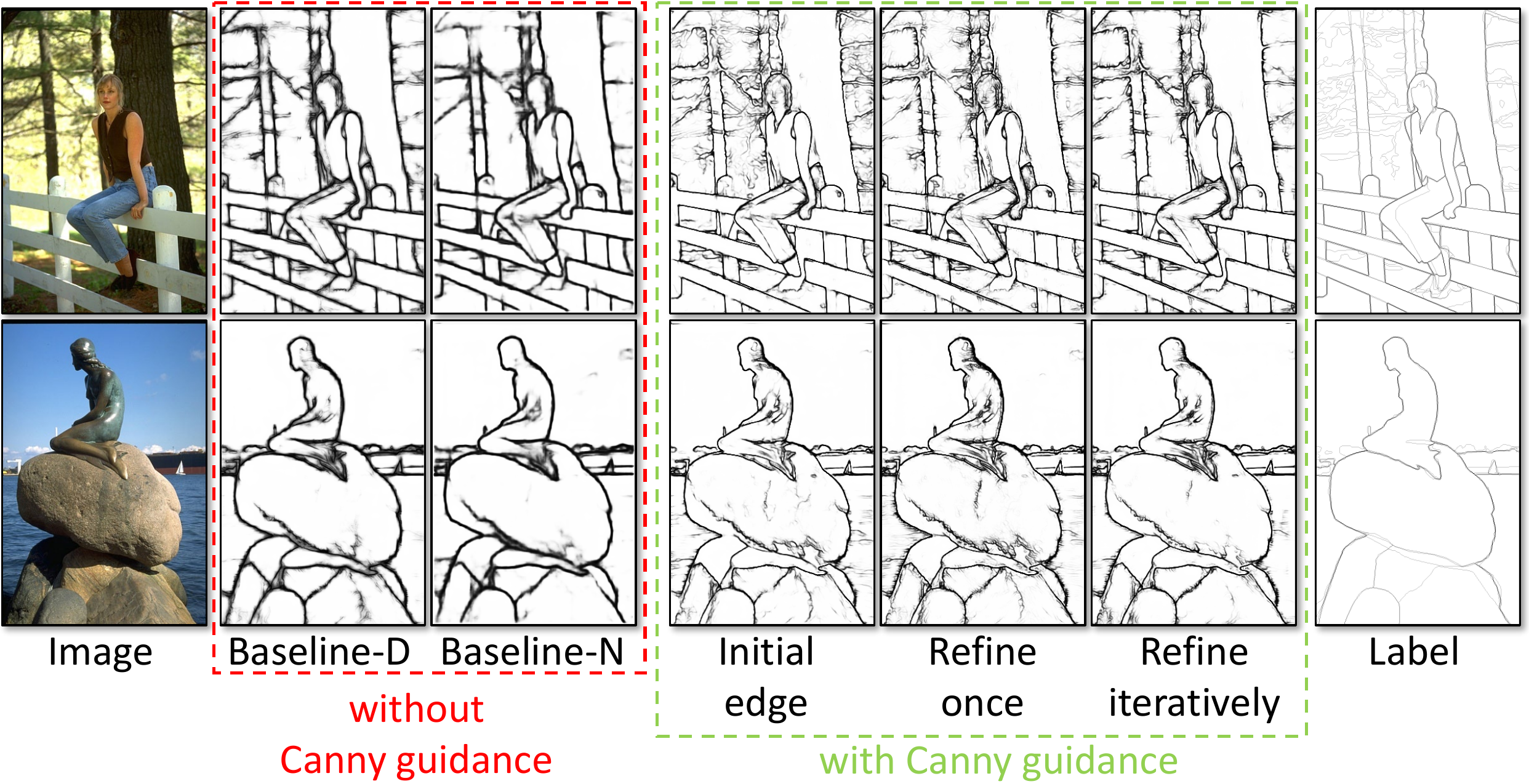}
			\caption{Ablations for various refinement strategies.}
			\label{fig:ablations}
		\end{figure}

		Moreover, to verify the importance of Canny guidance, with all other settings unchanged, we conduct experiments without the Canny-guided strategy by simply dilating human labels to replace the over-detected Canny edge maps (Baseline-D), and directly generate refined training labels using NMS on the averaged labels from multiple annotators (Baseline-N).
		Qualitative and quantitative comparisons are shown in Figure \ref{fig:ablations} and Table \ref{tab:ablations}, respectively. We observe that the crispness drops significantly in all cases without Canny guidance, since Canny edges can be treated as, or close to true edges without any offsets and are essential for crisp edge detection, as studied in Section~\ref{sec: noisy_labels}. 
		After Canny-guided refinement, the trained models can predict crisper and cleaner edges.
		
		\begin{table}[htbp]
			\centering
			\scalebox{1}{
				\begin{tabular}{c|ccc}
					\hline
					Method & ODS   & OIS   & AC \\
					\hline
					
					\multicolumn{4}{c}{without Canny guidance} \\
					\hline
					
					Baseline-D & 0.769 & 0.778 & \textbf{0.271} \\
					Baseline-N & \textbf{0.775} & \textbf{0.788} & 0.252 \\
					
					\hline
					\multicolumn{4}{c}{with Canny guidance} \\
					
					\hline
					Initial edge & 0.735 & 0.75  & 0.41 \\
					Initial edge + Refine once & 0.753 & 0.766 & 0.414 \\
					Patch-based refine iteratively & \textbf{0.771} & \textbf{0.782} & \textbf{0.424} \\
					\hline
				\end{tabular}
			}
			\caption{Quantitative results for various refinement strategies.}
			\label{tab:ablations}
		\end{table}
		
		\subsubsection{Ablations for robustness}\label{sec: ablation_dropout}
		Our iterative strategy starts the refinement with an initial edge generated by computing the Hadamard product of over-detected Canny edges and human labels. Therefore, it is natural to come up with a question that how robust such a seed could be. In most cases, human labels and the corresponding Canny edges can coincide within small regions, producing a relatively complete initial edge and benefiting subsequent refinement. However, there still could be extreme cases that the initial edge map is badly sparse caused by slight offsets across a large region on human labels.
		
\begin{figure}[h]
		\centering
		\includegraphics[width=\linewidth]{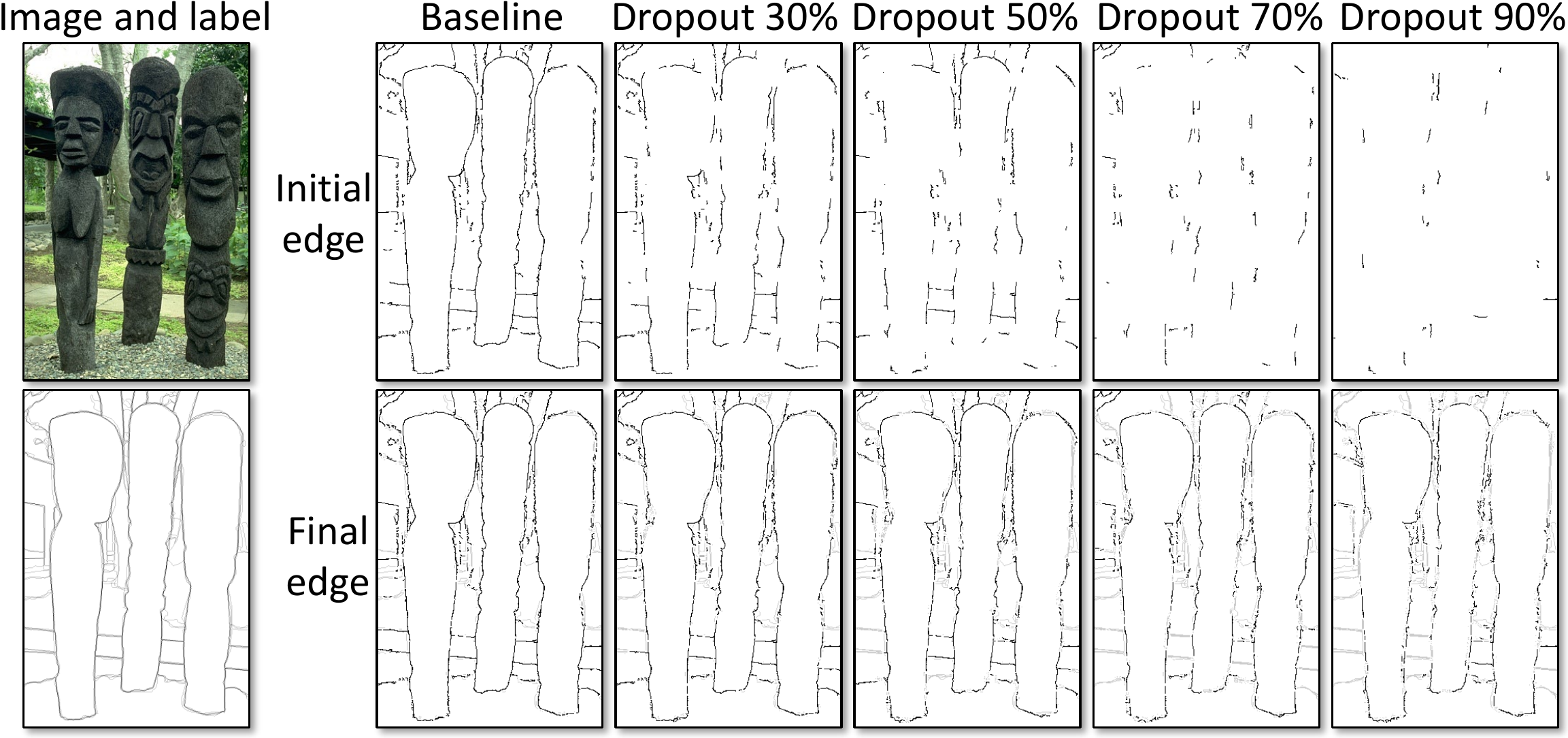}
		\caption{Ablations for the robustness of initial edges by dropout various levels of pixels (edges).}
		\label{fig:ablation_dropout}
	\end{figure}

		We conduct experiments to test the robustness of the initial edge maps by randomly dropout different percentages of pixels (edges) on them. The original initial edge maps without any dropout are treated as the baseline version. As demonstrated in Figure~\ref{fig:ablation_dropout}, when the dropout ratio is no more than 50\%, the recovered final edge map changes slightly compared with the original baseline, so as the quantitative performance change ($\sim$1\%) in Table~\ref{tab:ablations_dropout}; while if the initial edge maps are badly broken (dropout more than 70\% edges), some discontinuous regions become hard to be inpainted.
		However, in such cases, the human labels themselves are too noisy to work as ground truths for supervised methods. 

	\begin{table}[htbp]
		\centering
		\scalebox{1}{
			\begin{tabular}{c|ccc}
				\hline
				Method & ODS   & OIS & AC\\
				\hline
				
				Baseline & 0.771 & 0.782 &  0.424\\
				Dropout 30\% & 0.767 & 0.776 &  0.424\\
				Dropout 50\% & 0.762 & 0.771 &  0.427\\
				Dropout 70\% & 0.756 & 0.767 &  0.427\\
				Dropout 90\% & 0.742 & 0.750 &  0.431\\
				
				\hline
		\end{tabular}}
		\caption{Quantitative results of initial edges with different dropout levels.}
		\label{tab:ablations_dropout}
	\end{table}

	\subsubsection{Ablations for other crisp edge methods}
	We directly refine human labels to achieve crisp edge detection, which means our method can be easily integrated with other crisp edge detection works that focus on loss function designs~\cite{deng2018learning, huan2021unmixing}. 
	To thoroughly verify the compatibility of our method, we conduct experiments based on PiDiNet~\cite{su2021pixel} with Dice loss~\cite{deng2018learning} and tracing loss~\cite{huan2021unmixing} following their settings, training with original labels and our refined labels, and evaluating after and before NMS. 
	
\begin{figure*}[h]
		\centering
		\includegraphics[width=\linewidth]{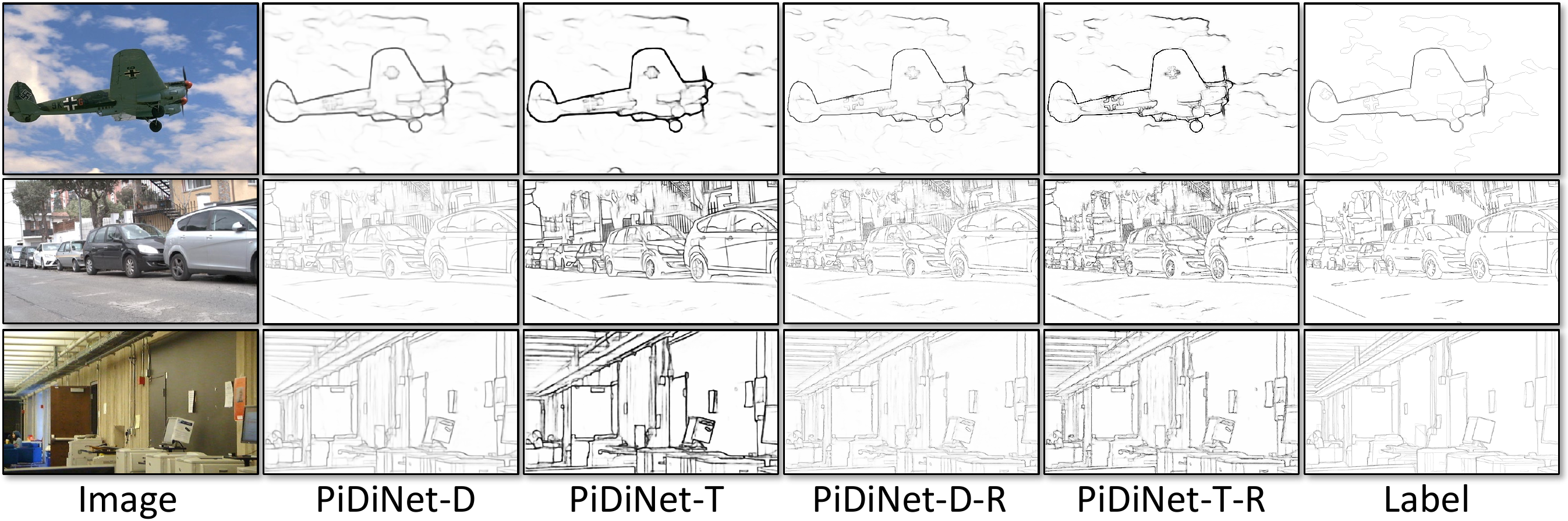}
		\caption{Qualitative results on datasets of BSDS500 (the first row), BIPED (the second row) and Multicue (the last row) before NMS. Although dice loss (``-D'') and tracing loss (``-T'') can generate crisp edges,  training with our refined labels (``-R'') can bring a free performance boost and make edge maps even crisper.}
		\label{fig:comparisons_dice_tracing}
	\end{figure*}

	By observing the qualitative and quantitative results in Figure~\ref{fig:comparisons_dice_tracing} and Table~\ref{tab:comparison_other_loss}, several interesting conclusions can be drawn. 
	(a) Compared with cross-entropy loss, Dice loss and tracing loss can generate crisper edge maps, but the improvement for AC is still less than only training with our refined labels. 
	(b) Dice loss and tracing loss can be perfectly integrated with our method and further improve AC in all cases.
	(c) Dice loss and tracing loss can enjoy a free performance boost after integrating our method in most cases. Especially, Dice loss training with refined labels can achieve the best performance, since dice loss emphasizes the image-level similarity between two sets of edges, which can balance the discontinuities in refined labels.
	(d) After integrating our method with Dice loss and tracing loss, the predicted edge maps are almost crisp enough that the performance change after and before NMS is very slight ($<$1\% in all cases). Such a phenomenon shows the potential to directly adopt the predicted results without NMS for downstream tasks, making the whole pipeline end-to-end.

	\begin{table}[htbp]
		\centering
		\scalebox{0.95}{
			\begin{tabular}{c|c|c|c|c|c|c}
				\hline
				\multirow{2}{*}{Dataset    } & \multirow{2}{*}{Methods} & \multicolumn{2}{c|}{after NMS} & \multicolumn{2}{c|}{before NMS} & \multirow{2}{*}{AC} \\
				\cline{3-6}          &       & ODS   & OIS   & ODS   & OIS   &  \\
				\hline
				\multirow{5}[2]{*}{BSDS 500} & PiDiNet-D & \textbf{0.782} & \textbf{0.798} & \textbf{0.774} & 0.788 & 0.306 \\
				& PiDiNet-T & 0.78  & 0.796 & 0.77  & 0.787 & 0.333 \\
				& PiDiNet-R & 0.771 & 0.782 & 0.771 & 0.78  & 0.424 \\
				& PiDiNet-D-R & 0.779 & 0.794 & 0.773 & 0.787 & \textbf{0.546} \\
				& PiDiNet-T-R & 0.773 & 0.79  & 0.772 & \textbf{0.789} & 0.545 \\
				\hline
				\multicolumn{1}{c|}{\multirow{5}[2]{*}{Multicue}} & PiDiNet-D & 0.871 & 0.876 & 0.77  & 0.779 & 0.208 \\
				& PiDiNet-T & 0.866 & 0.871 & 0.742 & 0.754 & 0.217 \\
				& PiDiNet-R & 0.892 & 0.907 & 0.886 & 0.904 & 0.424 \\
				& PiDiNet-D-R & \textbf{0.899} & \textbf{0.912} & \textbf{0.894} & \textbf{0.909} & 0.463 \\
				& PiDiNet-T-R & 0.884 & 0.903 & 0.876 & 0.894 & \textbf{0.559} \\
				\hline
				\multirow{5}[2]{*}{BIPED} & PiDiNet-D & \textbf{0.873} & 0.877 & 0.852 & 0.858 & 0.34 \\
				& PiDiNet-T & 0.872 & 0.878 & 0.845 & 0.852 & 0.296 \\
				& PiDiNet-R & 0.863 & 0.873 & 0.863 & 0.873 & 0.512 \\
				& PiDiNet-D-R & 0.869 & \textbf{0.879} & \textbf{0.867} & \textbf{0.877} & 0.696 \\
				& PiDiNet-T-R & 0.863 & 0.874 & 0.862 & 0.873 & \textbf{0.704} \\
				\hline
		\end{tabular}}
		\caption{Quantitative results for PiDiNet when training with original labels and refined labels based on dice loss and tracing loss. ``-D'' and ``-T'' means training with dice loss and tracing loss. ``-D-R'' and ``-T-R'' means training with our refined labels with dice loss and tracing loss, respectively.}
		\label{tab:comparison_other_loss}
	\end{table}%

	\subsection{Crisp Edge Detection for Other Vision Tasks}
	\subsubsection{Optical Flow}\label{sec:optical_flow}
	Optical flow estimation is the task of estimating per-pixel motion cues between consecutive video frames. The main difficulties include large displacements by fast-moving objects, occlusions, motion blur, and textureless surfaces. Based on the observation that motion boundaries often tend to appear at image edges, EpicFlow~\cite{revaud2015epicflow} computes sparse matches from DeepMatching~\cite{weinzaepfel2013deepflow} and leverages detected edges to perform sparse-to-dense interpolation relying on the edge-aware geodesic distance. The obtained dense correspondences thus are robust to large displacements and motion discontinuities. Due to the nature of the EpicFlow algorithm, the quality of edge detection is important to the final results, where crisper edges are much cleaner and bring less ambiguity. To show the advantage of crisp edge detection, we replace the edge detection part in EpicFlow with PiDiNet trained on the BSDS500 dataset with original labels and refined ones, respectively. Detailed settings can be seen in Section~\ref{sec:dataset}.
	
\begin{figure}[h]
		\centering
		\includegraphics[width=\linewidth]{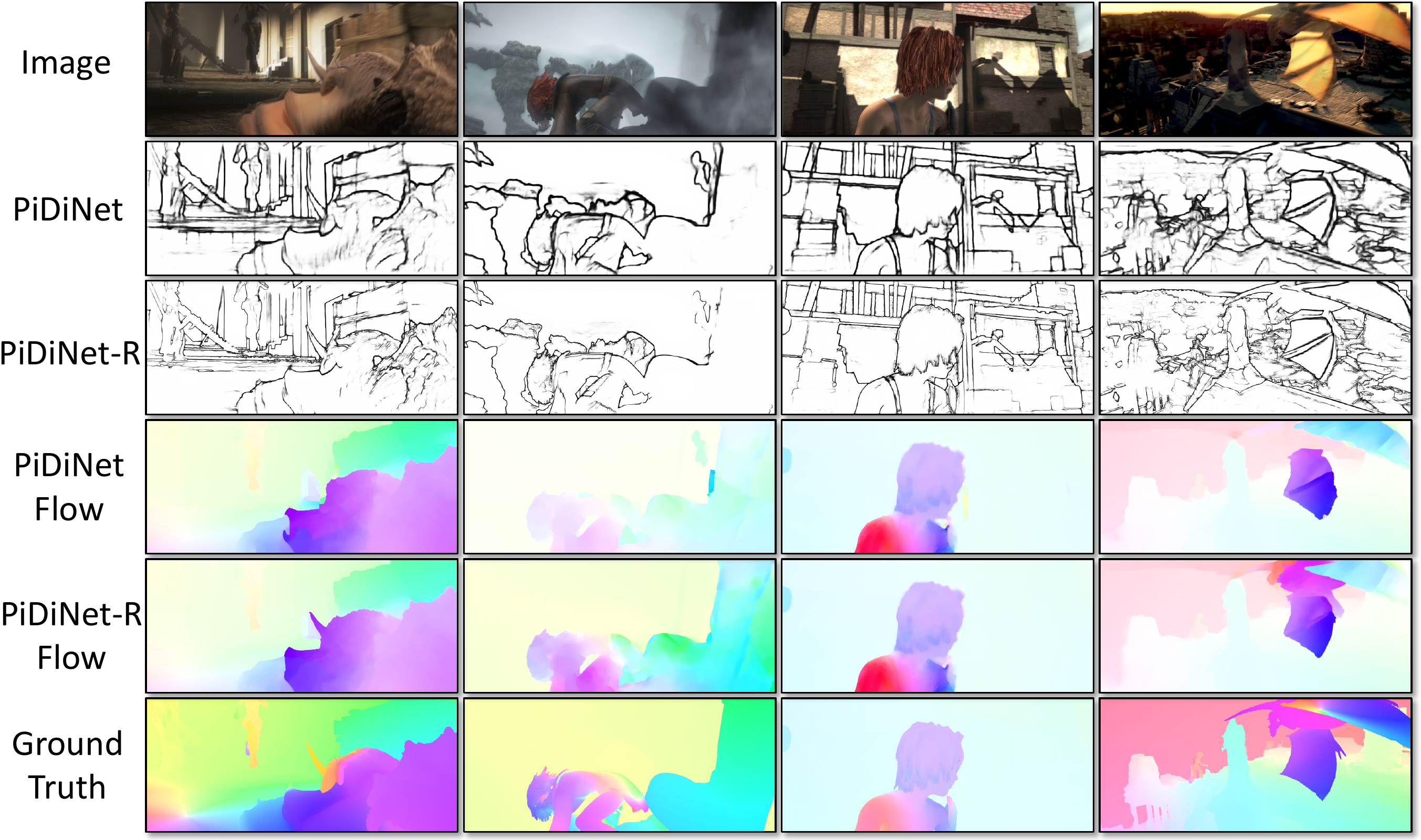}
		\caption{Illustration of optical flow estimation examples on Sintel dataset. Crisp edge maps bring better localization near motion discontinuities.}
		\label{fig:optical_flow_sintel}
	\end{figure}

\begin{figure}[h]
		\centering
		\includegraphics[width=\linewidth]{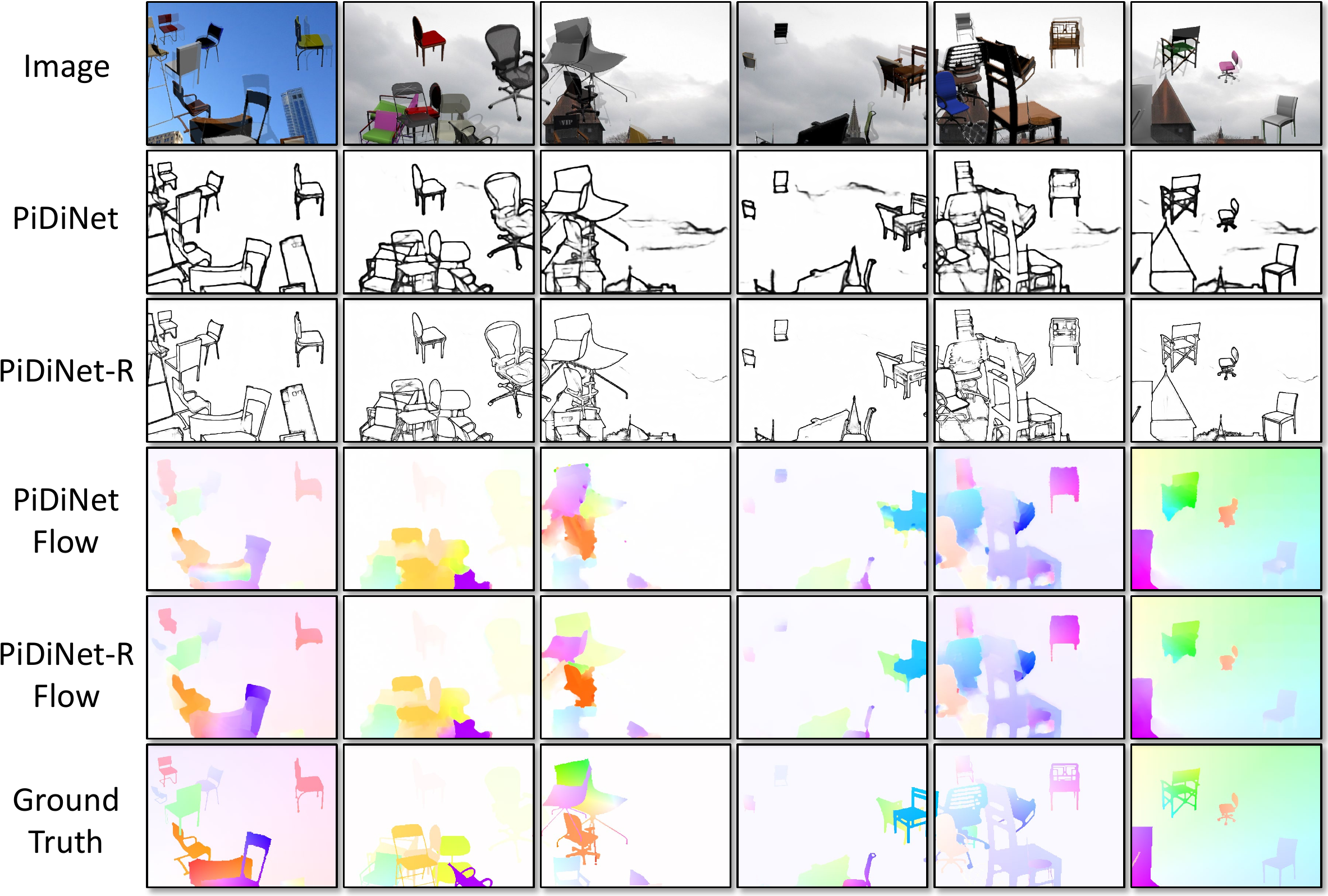}
		\caption{Illustration of optical flow estimation examples on FlyingChairs dataset. Crisp edge maps yield more accurate results with the same baseline algorithm.}
		\label{fig:optical_flow_flyingchairs}
	\end{figure}

	We adopt two widely-used optical flow datasets, Sintel~\cite{butler2012naturalistic} and FlyingChairs~\cite{dosovitskiy2015flownet} in this paper. For the Sintel dataset, we conduct experiments on the training set of the ``final'' version that features realistic rendering effects such as motion, defocus blur and atmospheric effects. For the FlyingChairs dataset, since the whole dataset is relatively large, we only take the first 1k samples, which is enough for evaluations. As in previous works, we adopt Averaged Endpoint Error (AEE) as the evaluation metric (lower is better). Experimental results are summarized in Table~\ref{tab:optical_flow}, where PiDiNet trained with our refined labels achieves better AEE performance than the original thick one. We also present some qualitative examples of the original and crisp edges together with their optical flow estimations in Figure~\ref{fig:optical_flow_sintel} and~\ref{fig:optical_flow_flyingchairs}. Without fine-tuning the parameters under the same baseline of EpicFlow~\cite{revaud2015epicflow}, we can observe that crisp edge maps reduce ambiguities and bring more accurate and clear results near object boundaries. Both qualitative and quantitative results show the advantage of crisp edge detection for optical flow estimation.

	\begin{table}[htbp]
		\centering
		\begin{tabular}{c|cc|cc}
			\hline
			& \multicolumn{2}{c|}{Sintel} & \multicolumn{2}{c}{FlyingChairs} \\
			\hline
			Methods & AEE   & AC    & AEE   & AC \\
			\hline
			Baseline EpicFlow~\cite{revaud2015epicflow} & 3.686 & \textbackslash{} & \textbackslash{} & \textbackslash{} \\
			PiDiNet-EpicFlow & 3.627 & 0.254 & 2.734 & 0.264 \\
			PiDiNet-R-EpicFlow & \textbf{3.602} & \textbf{0.445} & \textbf{2.718} & \textbf{0.448} \\
			\hline
		\end{tabular}
		\caption{Optical flow estimation performance achieved with PiDiNet edge detection integrated by the EpicFlow~\cite{revaud2015epicflow} on Sintel and FlyingChairs dataset. ``-R'' means training with refined labels.}
		\label{tab:optical_flow}%
	\end{table}%
	
	\subsubsection{Semantic Segmentation}
	Semantic segmentation is an important high-level vision task that aims at densely classifying each pixel of an image with the semantic category. It achieves a significant performance boost with the help of fully convolutional networks~\cite{long2015fully}. However, the successive down-sampling layers generate low-resolution feature maps and fail to capture precise boundaries of objects. To address this issue, Bertasius et al.~\cite{bertasius2016semantic} proposed Boundary Neural Field (BNF) to integrate the initial predictions with edge cues by minimizing the global boundary-based energy. Therefore, an accurate and crisp edge map is important for BNF. To show the effectiveness of crisp edge detection, we adopt DeepLabV3+~\cite{chen2018encoder} with various backbones (MobileNet~\cite{howard2017mobilenets} and ResNet50~\cite{he2016deep}) to generate the initial segmentation results, then apply our edges maps and BNF to further improve the predictions. Edge maps are generated by PiDiNet trained on original and refined labels the same as~\ref{sec:optical_flow}.
	
\begin{figure}[h]
		\centering
		\includegraphics[width=1.0\linewidth]{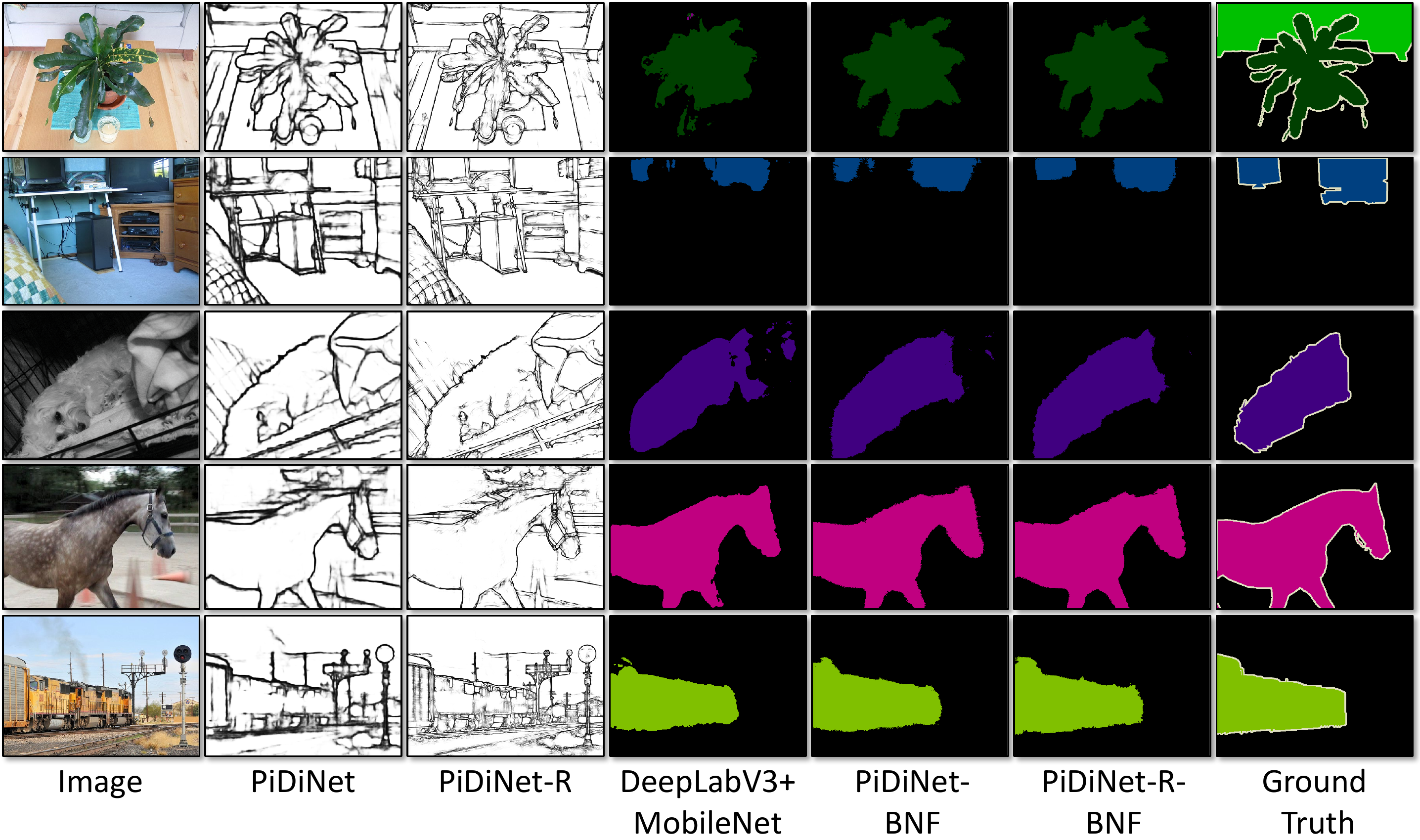}
		\caption{Illustration of semantic segmentation examples on Pascal Context validation set. Crisp edge maps can bring more complete segmentation and better localization near object boundaries.}
		\label{fig:semantic_segmentation_voc}
	\end{figure}

	We conduct experiments on the Pascal Context validation set~\cite{everingham2010pascal}, and report three widely-adopted segmentation evaluation metrics, pixel accuracy (PA), mean pixel accuracy (MPA) and mean intersection over union (mIOU). Quantitative results are summarized in Table~\ref{tab:segmentation}, where PiDiNet trained with refined labels achieves the best performance on all metrics with all backbones, which proves the advantages of crisp edge detection. We illustrate some qualitative examples of the original and crisp edges and their refined segmentation results in Figure~\ref{fig:semantic_segmentation_voc}. These results illustrate that semantic segmentation can benefit from better-localized edge maps, and be further improved by crisper ones that are better at capturing the precise contour of semantic objects.
	
	\begin{table}[htbp]
		\centering
		\begin{tabular}{c|cccc}
			\hline
			Method & PA    & MPA   & mIOU & AC \\
			\hline
			DeepLabV3+ MobileNet~\cite{howard2017mobilenets} & 0.918 & 0.779 & 0.666 & \textbackslash{} \\
			PiDiNet-BNF & 0.922 & 0.790  & 0.680 & 0.252 \\
			PiDiNet-R-BNF & \textbf{0.923} & \textbf{0.803} & \textbf{0.684} & \textbf{0.444} \\
			\hline
			DeepLabV3+ ResNet50~\cite{he2016deep} & 0.939 & 0.857 & 0.744 & \textbackslash{} \\
			PiDiNet-BNF & 0.939 & 0.862 & 0.747 & 0.252 \\
			PiDiNet-R-BNF & \textbf{0.940} & \textbf{0.871} & \textbf{0.749} & \textbf{0.444} \\
			\hline
		\end{tabular}
		\caption{Semantic segmentation performance achieved with PiDiNet edge detection integrated by BNF~\cite{bertasius2016semantic} on Pascal Context validation set. ``-R'' means training with refined labels. We report the initial predictions and their refined results on two backbones of MobileNet and ResNet50.}
		\label{tab:segmentation}
	\end{table}
	
	\section{Conclusions and Limitations}
	In this paper, we explore the reason of predicting ``thick'' edges by deep learning edge detectors. We introduce the first crispness metric for edge detection and explore the impact of noisy human-labeled edges. We find that the issue of thick predictions is mainly caused by noisy human labels, and the problem is aggravated by the cross-entropy loss. Based on the observation, we propose an iterative Canny-guided refinement method to refine human labels. 
	Our pipeline does not introduce any extra modules or loss functions in training and testing, and can be integrated with any edge detection backbones. 
	Comprehensive experiments demonstrate the effectiveness of our method for improving the edge crispness compared to models trained from original labels. We also verify the superiority of crisp edge detection for other downstream vision tasks such as optical flow estimation and semantic segmentation.
	
	\begin{figure}[h]
		\centering
		\includegraphics[width=1.0\linewidth]{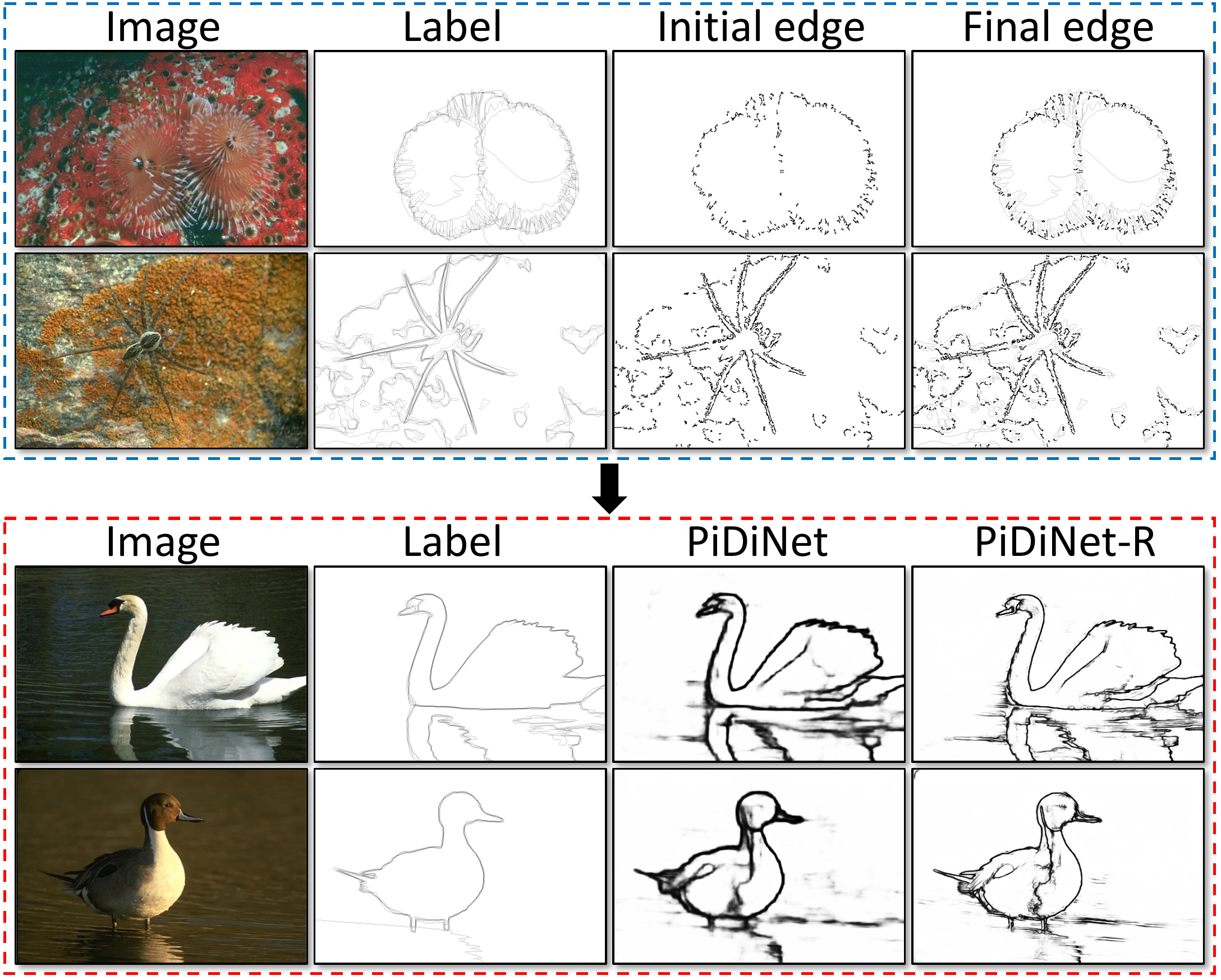}
		\caption{Examples of some failure cases. The refined training labels and predicted results are inside the blue and red frames, respectively.}
		\label{fig:failure_case}
	\end{figure}
	
	\textit{Limitations}: 
	Although training with refined labels can generate crisp edges, the performance of our plug-and-play method still relies on the adopted edge detector and the quality of human labels. As shown in Figure~\ref{fig:failure_case}, we present some failure cases of refined labels in the blue frame and prediction results in the red frame.
	
	As demonstrated in the blue frame, failure cases often happen on images with rich texture and unclear boundaries, which is also challenging for human annotators. In such cases, the produced initial edges will be so sparse that the discontinuities are hard to be inpainted without enough semantic context, leading to unsatisfactory final edges.
	
	Consequently, as illustrated in the red frame, due to the strong generalization ability of learning-based edge detection methods, the prediction results may still contain some salient noises (e.g. water reflection) that are not semantically meaningful. Our method may amplify such existing salient noises more depending on the performance of the edge detection backbone and its original predictions. To tackle this problem, training a better edge inpainting network can be an interesting direction for future work.

	\bibliographystyle{IEEEtran}
	\bibliography{references}

\begin{thebibliography}{10}
\providecommand{\url}[1]{#1}
\csname url@samestyle\endcsname
\providecommand{\newblock}{\relax}
\providecommand{\bibinfo}[2]{#2}
\providecommand{\BIBentrySTDinterwordspacing}{\spaceskip=0pt\relax}
\providecommand{\BIBentryALTinterwordstretchfactor}{4}
\providecommand{\BIBentryALTinterwordspacing}{\spaceskip=\fontdimen2\font plus
\BIBentryALTinterwordstretchfactor\fontdimen3\font minus
  \fontdimen4\font\relax}
\providecommand{\BIBforeignlanguage}[2]{{%
\expandafter\ifx\csname l@#1\endcsname\relax
\typeout{** WARNING: IEEEtran.bst: No hyphenation pattern has been}%
\typeout{** loaded for the language `#1'. Using the pattern for}%
\typeout{** the default language instead.}%
\else
\language=\csname l@#1\endcsname
\fi
#2}}
\providecommand{\BIBdecl}{\relax}
\BIBdecl

\bibitem{zitnick2014edge}
C.~L. Zitnick and P.~Doll{\'a}r, ``Edge boxes: Locating object proposals from
  edges,'' in \emph{European conference on computer vision}.\hskip 1em plus
  0.5em minus 0.4em\relax Springer, 2014, pp. 391--405.

\bibitem{arbelaez2014multiscale}
P.~Arbel{\'a}ez, J.~Pont-Tuset, J.~T. Barron, F.~Marques, and J.~Malik,
  ``Multiscale combinatorial grouping,'' in \emph{Proceedings of the IEEE
  conference on computer vision and pattern recognition}, 2014, pp. 328--335.

\bibitem{revaud2015epicflow}
J.~Revaud, P.~Weinzaepfel, Z.~Harchaoui, and C.~Schmid, ``Epicflow:
  Edge-preserving interpolation of correspondences for optical flow,'' in
  \emph{Proceedings of the IEEE conference on computer vision and pattern
  recognition}, 2015, pp. 1164--1172.

\bibitem{bertasius2016semantic}
G.~Bertasius, J.~Shi, and L.~Torresani, ``Semantic segmentation with boundary
  neural fields,'' in \emph{Proceedings of the IEEE conference on computer
  vision and pattern recognition}, 2016, pp. 3602--3610.

\bibitem{cheng2020boundary}
T.~Cheng, X.~Wang, L.~Huang, and W.~Liu, ``Boundary-preserving mask r-cnn,'' in
  \emph{European conference on computer vision}.\hskip 1em plus 0.5em minus
  0.4em\relax Springer, 2020, pp. 660--676.

\bibitem{xiong2019foreground}
W.~Xiong, J.~Yu, Z.~Lin, J.~Yang, X.~Lu, C.~Barnes, and J.~Luo,
  ``Foreground-aware image inpainting,'' in \emph{Proceedings of the IEEE/CVF
  Conference on Computer Vision and Pattern Recognition}, 2019, pp. 5840--5848.

\bibitem{nazeri2019edgeconnect}
K.~Nazeri, E.~Ng, T.~Joseph, F.~Z. Qureshi, and M.~Ebrahimi, ``Edgeconnect:
  Generative image inpainting with adversarial edge learning,'' \emph{arXiv
  preprint arXiv:1901.00212}, 2019.

\bibitem{kittler1983accuracy}
J.~Kittler, ``On the accuracy of the sobel edge detector,'' \emph{Image and
  Vision Computing}, vol.~1, no.~1, pp. 37--42, 1983.

\bibitem{canny1986computational}
J.~Canny, ``A computational approach to edge detection,'' \emph{IEEE
  Transactions on pattern analysis and machine intelligence}, no.~6, pp.
  679--698, 1986.

\bibitem{xie2015holistically}
S.~Xie and Z.~Tu, ``Holistically-nested edge detection,'' in \emph{Proceedings
  of the IEEE international conference on computer vision}, 2015, pp.
  1395--1403.

\bibitem{liu2017richer}
Y.~Liu, M.-M. Cheng, X.~Hu, K.~Wang, and X.~Bai, ``Richer convolutional
  features for edge detection,'' in \emph{Proceedings of the IEEE conference on
  computer vision and pattern recognition}, 2017, pp. 3000--3009.

\bibitem{he2019bi}
J.~He, S.~Zhang, M.~Yang, Y.~Shan, and T.~Huang, ``Bi-directional cascade
  network for perceptual edge detection,'' in \emph{Proceedings of the IEEE/CVF
  Conference on Computer Vision and Pattern Recognition}, 2019, pp. 3828--3837.

\bibitem{simonyan2014very}
K.~Simonyan and A.~Zisserman, ``Very deep convolutional networks for
  large-scale image recognition,'' \emph{arXiv preprint arXiv:1409.1556}, 2014.

\bibitem{wang2017deep}
Y.~Wang, X.~Zhao, and K.~Huang, ``Deep crisp boundaries,'' in \emph{Proceedings
  of the IEEE conference on computer vision and pattern recognition}, 2017, pp.
  3892--3900.

\bibitem{deng2018learning}
R.~Deng, C.~Shen, S.~Liu, H.~Wang, and X.~Liu, ``Learning to predict crisp
  boundaries,'' in \emph{Proceedings of the European Conference on Computer
  Vision (ECCV)}, 2018, pp. 562--578.

\bibitem{yu2018simultaneous}
Z.~Yu, W.~Liu, Y.~Zou, C.~Feng, S.~Ramalingam, B.~Kumar, and J.~Kautz,
  ``Simultaneous edge alignment and learning,'' in \emph{Proceedings of the
  European Conference on Computer Vision (ECCV)}, 2018, pp. 388--404.

\bibitem{acuna2019devil}
D.~Acuna, A.~Kar, and S.~Fidler, ``Devil is in the edges: Learning semantic
  boundaries from noisy annotations,'' in \emph{Proceedings of the IEEE/CVF
  Conference on Computer Vision and Pattern Recognition}, 2019, pp.
  11\,075--11\,083.

\bibitem{arbelaez2010contour}
P.~Arbelaez, M.~Maire, C.~Fowlkes, and J.~Malik, ``Contour detection and
  hierarchical image segmentation,'' \emph{IEEE transactions on pattern
  analysis and machine intelligence}, vol.~33, no.~5, pp. 898--916, 2010.

\bibitem{mely2016systematic}
D.~A. M{\'e}ly, J.~Kim, M.~McGill, Y.~Guo, and T.~Serre, ``A systematic
  comparison between visual cues for boundary detection,'' \emph{Vision
  research}, vol. 120, pp. 93--107, 2016.

\bibitem{poma2020dense}
X.~S. Poma, E.~Riba, and A.~Sappa, ``Dense extreme inception network: Towards a
  robust cnn model for edge detection,'' in \emph{Proceedings of the IEEE/CVF
  Winter Conference on Applications of Computer Vision}, 2020, pp. 1923--1932.

\bibitem{su2021pixel}
Z.~Su, W.~Liu, Z.~Yu, D.~Hu, Q.~Liao, Q.~Tian, M.~Pietik{\"a}inen, and L.~Liu,
  ``Pixel difference networks for efficient edge detection,'' in
  \emph{Proceedings of the IEEE/CVF International Conference on Computer
  Vision}, 2021, pp. 5117--5127.

\bibitem{dollar2014fast}
P.~Doll{\'a}r and C.~L. Zitnick, ``Fast edge detection using structured
  forests,'' \emph{IEEE transactions on pattern analysis and machine
  intelligence}, vol.~37, no.~8, pp. 1558--1570, 2014.

\bibitem{wibisono2020fined}
J.~K. Wibisono and H.-M. Hang, ``Fined: Fast inference network for edge
  detection,'' \emph{arXiv preprint arXiv:2012.08392}, 2020.

\bibitem{huan2021unmixing}
L.~Huan, N.~Xue, X.~Zheng, W.~He, J.~Gong, and G.-S. Xia, ``Unmixing
  convolutional features for crisp edge detection,'' \emph{IEEE Transactions on
  Pattern Analysis and Machine Intelligence}, 2021.

\bibitem{pathak2016context}
D.~Pathak, P.~Krahenbuhl, J.~Donahue, T.~Darrell, and A.~A. Efros, ``Context
  encoders: Feature learning by inpainting,'' in \emph{Proceedings of the IEEE
  conference on computer vision and pattern recognition}, 2016, pp. 2536--2544.

\bibitem{iizuka2017globally}
S.~Iizuka, E.~Simo-Serra, and H.~Ishikawa, ``Globally and locally consistent
  image completion,'' \emph{ACM Transactions on Graphics (ToG)}, vol.~36,
  no.~4, pp. 1--14, 2017.

\bibitem{yu2015multi}
F.~Yu and V.~Koltun, ``Multi-scale context aggregation by dilated
  convolutions,'' \emph{arXiv preprint arXiv:1511.07122}, 2015.

\bibitem{ren2019structureflow}
Y.~Ren, X.~Yu, R.~Zhang, T.~H. Li, S.~Liu, and G.~Li, ``Structureflow: Image
  inpainting via structure-aware appearance flow,'' in \emph{Proceedings of the
  IEEE/CVF International Conference on Computer Vision}, 2019, pp. 181--190.

\bibitem{shih20203d}
M.-L. Shih, S.-Y. Su, J.~Kopf, and J.-B. Huang, ``3d photography using
  context-aware layered depth inpainting,'' in \emph{Proceedings of the
  IEEE/CVF Conference on Computer Vision and Pattern Recognition}, 2020, pp.
  8028--8038.

\bibitem{huang2019indoor}
Y.-K. Huang, T.-H. Wu, Y.-C. Liu, and W.~H. Hsu, ``Indoor depth completion with
  boundary consistency and self-attention,'' in \emph{Proceedings of the
  IEEE/CVF International Conference on Computer Vision Workshops}, 2019, pp.
  0--0.

\bibitem{zhu2020edge}
S.~Zhu, G.~Brazil, and X.~Liu, ``The edge of depth: Explicit constraints
  between segmentation and depth,'' in \emph{Proceedings of the IEEE/CVF
  Conference on Computer Vision and Pattern Recognition}, 2020, pp.
  13\,116--13\,125.

\bibitem{song2020edgestereo}
X.~Song, X.~Zhao, L.~Fang, H.~Hu, and Y.~Yu, ``Edgestereo: An effective
  multi-task learning network for stereo matching and edge detection,''
  \emph{International Journal of Computer Vision}, vol. 128, no.~4, pp.
  910--930, 2020.

\bibitem{kirillov2017instancecut}
A.~Kirillov, E.~Levinkov, B.~Andres, B.~Savchynskyy, and C.~Rother,
  ``Instancecut: from edges to instances with multicut,'' in \emph{Proceedings
  of the IEEE Conference on Computer Vision and Pattern Recognition}, 2017, pp.
  5008--5017.

\bibitem{he2021enhanced}
H.~He, X.~Li, G.~Cheng, J.~Shi, Y.~Tong, G.~Meng, V.~Prinet, and L.~Weng,
  ``Enhanced boundary learning for glass-like object segmentation,'' in
  \emph{Proceedings of the IEEE/CVF International Conference on Computer
  Vision}, 2021, pp. 15\,859--15\,868.

\bibitem{yang2016object}
J.~Yang, B.~Price, S.~Cohen, H.~Lee, and M.-H. Yang, ``Object contour detection
  with a fully convolutional encoder-decoder network,'' in \emph{Proceedings of
  the IEEE conference on computer vision and pattern recognition}, 2016, pp.
  193--202.

\bibitem{simard2003best}
P.~Y. Simard, D.~Steinkraus, J.~C. Platt \emph{et~al.}, ``Best practices for
  convolutional neural networks applied to visual document analysis.'' in
  \emph{Icdar}, vol.~3, no. 2003.\hskip 1em plus 0.5em minus 0.4em\relax
  Edinburgh, 2003.

\bibitem{jackson2019style}
P.~T. Jackson, A.~A. Abarghouei, S.~Bonner, T.~P. Breckon, and B.~Obara,
  ``Style augmentation: data augmentation via style randomization.'' in
  \emph{CVPR workshops}, vol.~6, 2019, pp. 10--11.

\bibitem{zhao2020differentiable}
S.~Zhao, Z.~Liu, J.~Lin, J.-Y. Zhu, and S.~Han, ``Differentiable augmentation
  for data-efficient gan training,'' \emph{Advances in Neural Information
  Processing Systems}, vol.~33, pp. 7559--7570, 2020.

\bibitem{liu2018image}
G.~Liu, F.~A. Reda, K.~J. Shih, T.-C. Wang, A.~Tao, and B.~Catanzaro, ``Image
  inpainting for irregular holes using partial convolutions,'' in
  \emph{Proceedings of the European conference on computer vision (ECCV)},
  2018, pp. 85--100.

\bibitem{paszke2019pytorch}
A.~Paszke, S.~Gross, F.~Massa, A.~Lerer, J.~Bradbury, G.~Chanan, T.~Killeen,
  Z.~Lin, N.~Gimelshein, L.~Antiga \emph{et~al.}, ``Pytorch: An imperative
  style, high-performance deep learning library,'' \emph{Advances in neural
  information processing systems}, vol.~32, 2019.

\bibitem{kingma2014adam}
D.~P. Kingma and J.~Ba, ``Adam: A method for stochastic optimization,''
  \emph{arXiv preprint arXiv:1412.6980}, 2014.

\bibitem{zhou2017places}
B.~Zhou, A.~Lapedriza, A.~Khosla, A.~Oliva, and A.~Torralba, ``Places: A 10
  million image database for scene recognition,'' \emph{IEEE transactions on
  pattern analysis and machine intelligence}, vol.~40, no.~6, pp. 1452--1464,
  2017.

\bibitem{weinzaepfel2013deepflow}
P.~Weinzaepfel, J.~Revaud, Z.~Harchaoui, and C.~Schmid, ``Deepflow: Large
  displacement optical flow with deep matching,'' in \emph{Proceedings of the
  IEEE international conference on computer vision}, 2013, pp. 1385--1392.

\bibitem{butler2012naturalistic}
D.~J. Butler, J.~Wulff, G.~B. Stanley, and M.~J. Black, ``A naturalistic open
  source movie for optical flow evaluation,'' in \emph{European conference on
  computer vision}.\hskip 1em plus 0.5em minus 0.4em\relax Springer, 2012, pp.
  611--625.

\bibitem{dosovitskiy2015flownet}
A.~Dosovitskiy, P.~Fischer, E.~Ilg, P.~Hausser, C.~Hazirbas, V.~Golkov, P.~Van
  Der~Smagt, D.~Cremers, and T.~Brox, ``Flownet: Learning optical flow with
  convolutional networks,'' in \emph{Proceedings of the IEEE international
  conference on computer vision}, 2015, pp. 2758--2766.

\bibitem{long2015fully}
J.~Long, E.~Shelhamer, and T.~Darrell, ``Fully convolutional networks for
  semantic segmentation,'' in \emph{Proceedings of the IEEE conference on
  computer vision and pattern recognition}, 2015, pp. 3431--3440.

\bibitem{chen2018encoder}
L.-C. Chen, Y.~Zhu, G.~Papandreou, F.~Schroff, and H.~Adam, ``Encoder-decoder
  with atrous separable convolution for semantic image segmentation,'' in
  \emph{Proceedings of the European conference on computer vision (ECCV)},
  2018, pp. 801--818.

\bibitem{howard2017mobilenets}
A.~G. Howard, M.~Zhu, B.~Chen, D.~Kalenichenko, W.~Wang, T.~Weyand,
  M.~Andreetto, and H.~Adam, ``Mobilenets: Efficient convolutional neural
  networks for mobile vision applications,'' \emph{arXiv preprint
  arXiv:1704.04861}, 2017.

\bibitem{he2016deep}
K.~He, X.~Zhang, S.~Ren, and J.~Sun, ``Deep residual learning for image
  recognition,'' in \emph{Proceedings of the IEEE conference on computer vision
  and pattern recognition}, 2016, pp. 770--778.

\bibitem{everingham2010pascal}
M.~Everingham, L.~Van~Gool, C.~K. Williams, J.~Winn, and A.~Zisserman, ``The
  pascal visual object classes (voc) challenge,'' \emph{International journal
  of computer vision}, vol.~88, no.~2, pp. 303--338, 2010.

\end{thebibliography}

	
	%
	%
	%
	%

	\vfill
	
\end{document}